\documentclass[11pt, a4paper, logo, copyright]{googlecloud}

\pdftrailerid{redacted}

\makeatletter
\renewcommand\bibentry[1]{\nocitep{#1}{\frenchspacing\@nameuse{BR@r@#1\@extra@b@citeb}}}
\makeatother

\usepackage{kantlipsum, lipsum}
\usepackage{dsfont}
\usepackage[authoryear, sort&compress, round]{natbib}

\usepackage[utf8]{inputenc} %
\usepackage[T1]{fontenc}    %
\usepackage{hyperref}       %
\hypersetup{
  colorlinks,
  citecolor=blue,
  linkcolor=red,
  urlcolor=blue}
\usepackage{booktabs}       %
\usepackage{nicefrac}       %
\usepackage{microtype}      %
\usepackage{wrapfig} %
\usepackage{soul} %
\usepackage{tikz, adjustbox}
\usepackage{arydshln}
\usepackage[capitalize,noabbrev]{cleveref}
\usepackage{fancyvrb}

\usepackage{inconsolata}
\usepackage[skins,breakable,most]{tcolorbox}

\usepackage{caption}
\usepackage{subcaption}
\usepackage{enumitem}
\usepackage{flushend}
\usepackage{balance}
\usepackage{lineno}
\usepackage{amsmath,amssymb,amsfonts}
\usepackage{algorithm}
\usepackage{algpseudocode}
\usepackage{graphicx}
\usepackage{textcomp}
\usepackage{xcolor}
\usepackage{colortbl}
\usepackage{amsthm}
\usepackage{url}
\usepackage{float}
\usepackage{multirow}
\usepackage{multicol}
\usepackage{color}
\usepackage{bm}
\usepackage{bbm}

\usepackage{pifont}

\usepackage{array}
\newcolumntype{L}[1]{>{\raggedright\let\newline\\\arraybackslash\hspace{0pt}}m{#1}}
\newcolumntype{C}[1]{>{\centering\let\newline  \\\arraybackslash\hspace{0pt}}m{#1}}
\newcolumntype{R}[1]{>{\raggedleft\let\newline \\\arraybackslash\hspace{0pt}}m{#1}}

\usepackage[utf8]{inputenc} 
\usepackage[T1]{fontenc}    
\usepackage{amsfonts}       
\usepackage{comment} 
\usepackage{mdframed}
\usepackage{fontawesome}
\usepackage{csquotes}
\usepackage{makecell}
\definecolor{beigecolor}{RGB}{253, 244, 204} 
\definecolor{greencolor}{RGB}{228, 242, 217} 
\definecolor{bluecolor}{RGB}{66, 133, 244} 
\definecolor{orgcolor}{RGB}{255, 140, 15} 

\definecolor{redcolor}{RGB}{234, 67, 53} 

\definecolor{ggreen}{RGB}{52, 168, 83}

\definecolor{gyellow}{RGB}{251, 188, 5}

\lstdefinestyle{mystyle}{
    backgroundcolor=\color{backcolour},   
    commentstyle=\color{codegreen},
    keywordstyle=\color{magenta},
    numberstyle=\tiny\color{codegray},
    stringstyle=\color{codepurple},
    basicstyle=\ttfamily\scriptsize,
    breakatwhitespace=false,         
    breaklines=true,                 
    captionpos=b,                    
    keepspaces=true,                 
    numbers=left,                    
    numbersep=5pt,                  
    showspaces=false,                
    showstringspaces=false,
    showtabs=false,                  
    tabsize=2,
    frame=none,
    aboveskip=1pt,
    belowskip=1pt,
}
\lstdefinestyle{plainins}{
    backgroundcolor=\color{white},   
    commentstyle=\color{codegreen},
    keywordstyle=\color{magenta},
    numberstyle=\tiny\color{codegray},
    stringstyle=\color{codepurple},
    basicstyle=\ttfamily\scriptsize,
    breakatwhitespace=false,         
    breaklines=true,                 
    captionpos=b,                    
    keepspaces=true,                 
    numbers=none,                    
    numbersep=5pt,                  
    showspaces=false,                
    showstringspaces=false,
    showtabs=false,                  
    tabsize=2,
    aboveskip=0pt,
    belowskip=0pt,
    frame=single
}
\lstdefinestyle{plainexam}{
    backgroundcolor=\color[HTML]{FFFCF3},   
    commentstyle=\color{codegreen},
    keywordstyle=\color{magenta},
    numberstyle=\tiny\color{codegray},
    stringstyle=\color{codepurple},
    basicstyle=\ttfamily\scriptsize,
    breakatwhitespace=false,         
    breaklines=true,                 
    captionpos=b,                    
    keepspaces=true,                 
    numbers=none,                    
    numbersep=5pt,                  
    showspaces=false,                
    showstringspaces=false,
    showtabs=false,                  
    tabsize=2,
    aboveskip=0pt,
    belowskip=0pt
}

\tcbset{
  aibox/.style={
    width=\linewidth,
    top=5pt,
    bottom=1pt,
    colback=blue!6!white,
    colframe=black,
    colbacktitle=black,
    enhanced,
    center,
    fontupper=\scriptsize,
    attach boxed title to top left={yshift=-0.1in,xshift=0.15in},
    boxed title style={boxrule=0pt,colframe=white,},
  }
}
\definecolor{lightblue}{rgb}{0.22,0.45,0.70}

\newtcolorbox{AIbox}[2][]{aibox,title=#2,#1}

\title{SAGE: Steerable Agentic Data Generation for Deep Search
with Execution Feedback}
\renewcommand{\today}{}

\author[2]{Fangyuan Xu\textsuperscript{*}}
\author[1]{Rujun Han}
\author[1]{Yanfei Chen}
\author[1]{Zifeng Wang}
\author[1]{I-Hung Hsu}
\author[1]{Jun Yan}
\author[1]{Vishy Tirumalashetty}
\author[2]{Eunsol Choi}
\author[1]{Tomas Pfister}
\author[1]{Chen-Yu Lee}

\affil[1]{Google Cloud AI Research}
\affil[2]{New York University}

\begin{abstract}

Deep search agents, which aim to answer complex questions requiring reasoning across multiple documents, can significantly speed up the information-seeking process. Collecting human annotations for this application is prohibitively expensive due to long and complex exploration trajectories. 
We propose an agentic pipeline that automatically generates high-quality, difficulty-controlled deep search question-answer pairs for a given corpus and a target difficulty level. Our pipeline, \texttt{SAGE}, consists of a data generator which proposes QA pairs and a search agent which attempts to solve the generated question and provide execution feedback for the data generator. The two components interact over multiple rounds to iteratively refine the question-answer pairs until they satisfy the target difficulty level. Our intrinsic evaluation shows \texttt{SAGE} generates questions that require diverse reasoning strategies, while significantly increases the correctness and difficulty of the generated data. Our extrinsic evaluation demonstrates up to 23\% relative performance gain on popular deep search benchmarks by training deep search agents with our synthetic data. Additional experiments show that agents trained on our data can adapt from fixed-corpus retrieval to Google Search at inference time, without further training.

\end{abstract}

\begin{document}
\maketitle

\section{Introduction}

\begin{figure*}[t]
    \centering
    \includegraphics[scale=0.65,trim=10mm 0mm 0mm 0mm]{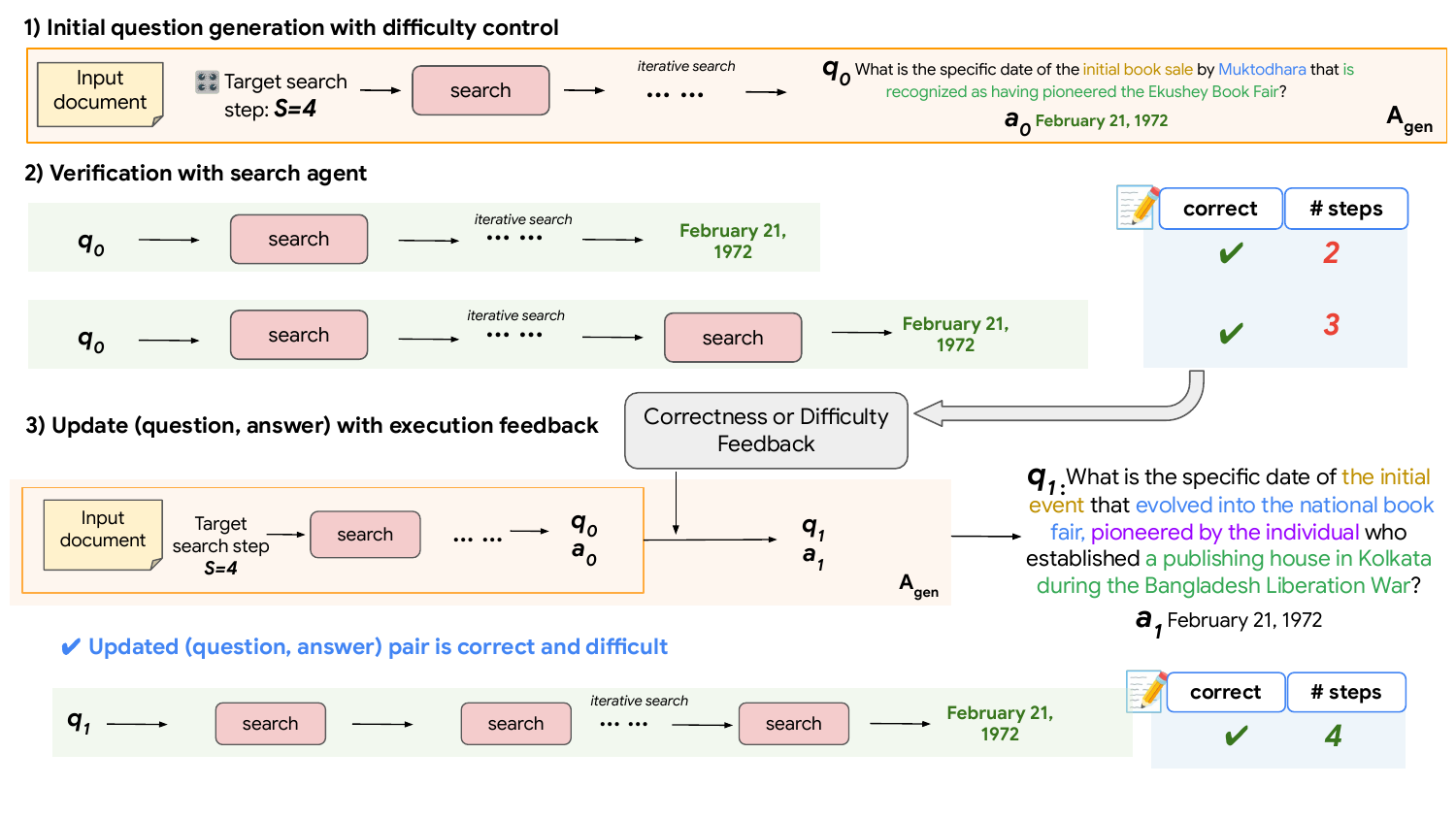}
    \caption{Illustration of our proposed method: (1) we first employ a data generator ($A_{gen}$) to generate an initial question, answer (QA) pair with a target number of search steps $S$ for difficulty control. Our framework then (2) collect execution traces from search agent ($A_{search}$) and (3) provide the execution feedback for the data generator agent to re-generate a QA pair. In this example, the initial QA pair generated does not satisfy the target step S=4. After re-generation with search agent's feedback, the new QA pair is both correct and difficult. Noticeably, the question complexity increases from $\mathbf{q_0}$ to $\mathbf{q_1}$.} 
    \label{fig:fig1}
\end{figure*}

Large language models (LLMs) are increasingly used as agents which interact with  external environments and solve complicated tasks, such as coding~\citep{jimenez2024swebench, surveycodeagent}, e-commerce and social forum discussion~\citep{zhou2024webarena, webmall}. Recently, there is a growing interest in building search-augmented agents that retrieve and reason about external information to solve complicated questions~\citep{trivedi-etal-2023-interleaving, asai2024selfrag, jin2025search}. High-quality, complex question–answer pairs are pivotal for training and evaluating capable search agents. Yet, such high quality data is not easily accessible and costly for humans to annotate~\citep{Wei2025BrowseCompAS, krishna-etal-2025-fact}.

Earlier work on retrieval-augmented generation (RAG) primarily focuses on questions where one search suffices to provide the necessary context~\citep{JoshiTriviaQA2017, kwiatkowski-etal-2019-natural, han2024ragqaarena}. Subsequent datasets~\citep{yang-etal-2018-hotpotqa,trivedi-etal-2022-musique} extended this setting to multi-hop reasoning. However, questions in these benchmarks typically require no more than four retrieval and reasoning steps. Furthermore, their dependence on extensive human annotation or pre-existing structural information (e.g. inter-document links) makes such approach difficult to scale to tasks that demand longer reasoning and search chains, limiting their applicability.

To address the limitations of prior work, we propose an agentic data generation pipeline that leverages search-augmented LLMs to generate high-quality, challenging data for training and evaluating deep search agents. In contrast to most existing search-augmented QA datasets, which begin with a question and then retrieve supporting evidence to find the answer~\citep{searchqa, kwiatkowski-etal-2019-natural}, our pipeline adopts reverse formulation to ensure the faithfulness of the generated data~\citep{Wei2025BrowseCompAS}. Specifically, given a randomly sampled document from a corpus and a target difficulty level, our pipeline employs a search-augmented data generator to iteratively search and reason, generating an initial question–answer(QA) pair grounded in the retrieved evidences.

We find that reverse generation alone is not sufficient to consistently produce correct QA pairs that require a targeted number of search steps. As the target number of steps increases, the data generator increasingly fails to generate correct QA pairs that satisfy the target difficulty level. This reveals a common pitfall: a mismatch between the data generator's intended search plans and the actual number of search steps required to answer the question, as illustrated in Figure \ref{fig:fig1}. Our observations inspire us to propose a dual-agent framework, which consists of a data generator agent and a search agent. The data generator agent first generates an initial draft of the question and answer pair. The generated question-answer pair is validated by a search agent to solve the question. We collect multiple traces from the search agent to gather feedback of the correctness and difficulty of the generated data. These execution traces are fed into the data generator to produce a new QA pair. 

Experiment results show that our framework, \texttt{SAGE} (\textbf{S}teerable \textbf{A}gentic \textbf{G}eneration with \textbf{E}xecution Feedback), significantly increases the quality of the generated data as measured by correctness and difficulty. We further conduct experiments on training search agents with our generated data, showing 27\% relative improvements on in-domain evaluation set and up to 23\% relative improvement for out-of-domain evaluation data~\citep{trivedi-etal-2022-musique, krishna-etal-2025-fact} across two model sizes (3B and 7B). Our analysis reveals that the generated data requires more diverse reasoning types than existing benchmarks.

\section{Background}

\subsection{Search agent}
Given an input question $\mathbf{q}$, a search agent $\mathcal{A}_{search}$ issues search queries ($s_0$, $s_1$, ... $s_i$) to a retrieval tool $\mathcal{T}$ to gather information in a multi-turn manner before providing the final answer $\mathbf{a}$. Following the ReACT~\citep{yao2023react} framework, the agent alternates between outputting reasoning traces $s_{i}$ and issuing search queries $s_{i}$,  outputting a sequence of $\{r_0, s_0, r_1, s_1, ... r_i, s_i, \mathbf{a}\}$. We omit the information returned by the search tool in the notation, which will be appended as input to the LM after $s_{i}$.

\subsection{Training a search agent}
While earlier work~\citep{trivedi-etal-2023-interleaving, yao2023react, Li2025Searcho1AS} explored prompting an LLM to use a search tool, recent work has shown that further training the model with either supervised learning~\citep{asai2024selfrag} or reinforcement learning~\citep{jin2025search, Nakano2021WebGPTBQ} is beneficial. Supervised fine-tuning requires a set of question $\mathbf{q}*$, answer $\mathbf{a}*$, as well as gold trajectory which consists of sub-queries, reasoning trace and information returned from search tool: $\{r_0, s_0, d_0, r_1, s_1, d_1 ... r_i, s_i, d_i\}$. As it is expensive to collect gold trajectories~\citep{Nakano2021WebGPTBQ, asai2024selfrag}, recent work~\citep{jin2025search} proposes training a search agent using reinforcement learning with outcome reward that measures the quality of the final answer $\mathbf{a}$ via exact match with a gold answer $\mathbf{a}*$. While training the model with reinforcement learning eliminates the need for gold trajectories, a set of high quality ($q$, $a$) pairs are essential for training a capable search agent.

\subsection{Existing training and evaluation data}

\begin{table}
\begin{center}
\footnotesize
\begin{tabular}{@{}llrr@{}}
\toprule
\textbf{Dataset} &  \textbf{Annotation}  & \textbf{\# search $\uparrow$} & \textbf{Avg@8 $\downarrow$}  \\ 
\midrule
\multicolumn{4}{c}{\textit{Large-scale training data}} \\
NQ &   Human   & 1.3  & 83.1 \\
HotpotQA  & Automatic &  2.1  & 82.9 \\
Musique & Automatic + human  & 2.7  & 64.4 \\
\multicolumn{4}{c}{\textit{Small-scale evaluation data}} \\
FRAMES  & Human & 3.2 & 74.3 \\
\midrule
\texttt{SAGE} & Automatic &  \textbf{4.9} & \textbf{79.5} \\
\bottomrule
\end{tabular} 
\end{center}
\caption{Comparison with other multi-hop question answering datasets. To estimate dataset difficulty, we report both average number of searches (\textbf{\# search}) and average performance of 8 samples from search-augmented gemini-2.5-flash.}
\label{tab:existing_datasets}
\end{table}

We review widely adopted datasets for training a search agent in Table~\ref{tab:existing_datasets}. While there are large-scale data annotated by human, they primarily consist of questions requiring a small number of steps to answer, as indicated by \textbf{\# search} in Table~\ref{tab:existing_datasets}. HotpotQA~\citep{yang-etal-2018-hotpotqa} and Musique~\citep{trivedi-etal-2022-musique} consist of multi-hop data, constructed through automatic or partially automatic pipelines. As constructing such a dataset can be cognitively heavy and time consuming for \textit{human}, recent effort focuses on relatively small-scale human annotated evaluation set to benchmark performances of LLM agents~\citep{krishna-etal-2025-fact, Wei2025BrowseCompAS}. 

Concurrent work such as WebDancer \citep{webdancer} and WebShaper \citep{webshaper} propose automatic training data construction pipeline. They focus on the setting of using a browsing tool as the retriever, which is non-trivial to reproduce \citep{Chen2025BrowseCompPlusAM} and expensive to train due to excessive API costs. Moreover, their large-scale data is not yet publicly available. We provide a more comprehensive discussion of prior and concurrent efforts to synthetically generate deep search training data in Section \ref{sec:related_work}.

\section{Generating Synthetic Data for Deep Search Agent: \texttt{SAGE}}

Our goal is to generate (question, answer) pairs that require a search agent to issue multiple calls to a search tool and reason over the retrieved information before deriving the final (question, answer) pair. Our framework consists of the following components: a data generator agent $\mathcal{A}_{gen}$, a search agent $\mathcal{A}_{search}$, a retrieval model $\mathcal{T}$ and a corpus $\mathcal{D}$ containing a set of documents. In this section, we describe each component of our framework. The full procedure is outlined in Algorithm~\ref{alg:data_generation} and we include the implementation details such as prompt, decoding setting in Section \ref{app:implementation_details} in the appendix.

\begin{algorithm*}[h]
\footnotesize
\begin{algorithmic}[1]
\Require $\mathcal{A}_{gen}$: data generator agent; $\mathcal{A}_{search}$: search agent; $d$: input document; $S$: target search steps; $K$: number of search traces; $R$: max feedback rounds.

\State $\mathcal{T}_{gen} \gets \emptyset, \mathcal{T}_{search} \gets \emptyset$ \Comment{Initialize accumulated $\mathcal{A}_{gen}$ and $\mathcal{A}_{search}$ traces.}
\State $\texttt{IS\_CORRECT} \gets False, \texttt{IS\_DIFFICULT} \gets False$

\For{$r \in \{0,\dots,R\}$} \label{alg:feedback_start}
    \If{$\texttt{IS\_CORRECT} \wedge \texttt{IS\_DIFFICULT}$}
        \State \textbf{break} 
    \Else

        \If{$r = 0$}
            \State $(q, a, t_{gen}) \gets \mathcal{A}_{gen}(d, S)$  \Comment{Generate initial (question, answer) pair} \label{alg:data_gen}
        \Else 
            \State $(q, a, t_{gen}) \gets \mathcal{A}_{gen}(q, a, S, \mathcal{T}_{gen}, \mathcal{T}_{search})$ \Comment{Re-generate the question and answer with execution feedback}
        \EndIf
        \State $(a^\star, S^\star, t_{search}, \texttt{IS\_CORRECT}, \texttt{IS\_DIFFICULT}) \gets \Call{RunSearchAgent}{q, a, S, K, \mathcal{A}_{search}}$ \Comment{Collect execution.}\label{alg:execution}
        \State $\mathcal{T}_{gen} \gets \mathcal{T}_{gen} \cup \{t_{gen}\}$, $\mathcal{T}_{search} \gets \mathcal{T}_{search} \cup \{t_{search}\}$ \Comment{Append generator and search traces.}
    \EndIf
\EndFor  \label{alg:feedback_end}
\If{$\texttt{IS\_CORRECT}$}
    \State \Return $(q, a)$ \Comment{Return correct pairs only.}
\EndIf

\end{algorithmic}
\caption{Agentic Data Generation with Execution Feedback}\label{alg:data_generation}
\end{algorithm*}

\begin{algorithm}[h]
\begin{algorithmic}[1]
\footnotesize
\Function{RunSearchAgent}{$q, a, S, K, \mathcal{A}_{search}$}
    \State $\mathcal{T} \gets [\ ]$ \Comment{All traces $(a_k', S_k')$}
    \State $\mathcal{M} \gets [\ ]$ \Comment{Correct traces where $a_k' = a$}
    \For{$k \in \{1,\dots,K\}$}
        \State $(a_k', S_k', t_k') \gets \mathcal{A}_{search}(q)$
        \State \texttt{append} $(a_k', S_k', t_k')$ to $\mathcal{T}$
        \State \texttt{append} $(a_k', S_k', t_k')$ to $\mathcal{M}$ \textbf{if} $a_k' = a$
    \EndFor
    \State $\texttt{IS\_CORRECT} \gets (|\mathcal{M}| > 0)$
    \If{$\texttt{IS\_CORRECT}$}
        \State $(a^\star, S^\star, t^\star) \gets \arg\min_{(a',S', t') \in \mathcal{M}} |S'|$ \Comment{Pick correct trace with least number of steps}
    \Else
        \State $(a^\star, S^\star, t^\star) \gets \textsc{UniformSample}(\mathcal{T})$ \Comment{Otherwise pick a random trace}

    \EndIf
    \State $\texttt{IS\_DIFFICULT} \gets (|S^\star| \ge S)$
    \State \Return $(a^\star, S^\star, t^\star, \texttt{IS\_CORRECT}, \texttt{IS\_DIFFICULT})$
\EndFunction

\end{algorithmic}
\caption{Verification with search agent}\label{alg:search_agent_verification}
\end{algorithm}
\vspace{-0.5em}

\subsection{Initial data generation with difficulty prompt}
Given an input document $d$ randomly sampled from a corpus $D$, a data generator agent is instructed to generate an initial (question, answer) pair $\{q, a\}$ by iteratively issuing  search query  ($s_0$, $s_1$, ... $s_i$) interleaved with reasoning traces to collect relevant information from the corpus (Line \ref{alg:data_gen} in Alg~\ref{alg:data_generation}). The iterative process ends when the agent outputs an initial $\{q, a\}$ pair.
The exact prompt used for this process is in Figure \ref{fig:initial_data_generator_prompt} in appendix. 

\paragraph{Difficulty prompt.} Questions can have various levels of difficulty depending on the number of search and reasoning steps. We propose to use the number of steps $S$ required by the search agent as a proxy measure for difficulty. Specifically, we include the target search step $S$ in the input prompt to the data generator agent and instruct it to reason and plan in order to produce a question that requires the target number of search steps to solve.

\subsection{Verifying the generated data}
Generated data can be incorrect or require less than the specified number of search steps to solve. To verify the quality of the generated data, we use a search agent $\mathcal{A}_{search}$ to solve the generated question (Line \ref{alg:execution} in Alg~\ref{alg:data_generation}). Given the generated question $q$, the search agent iteratively issues search queries before producing an answer $a'$. As the search agent itself is imperfect and may fail to find a solution for a valid $(q, a)$ pair, we sample $K$ traces from the search agent. We focus on two criteria for data quality: 
\begin{itemize}
    \item {\textbf{Correctness:} which measures whether the generated ($q$, $a$) pair is correct. We treat pass@K performance of answer produced by the search agent ($a'$) against the generated answer $a$ as correctness, following previous work~\citep{Shi2025PanguDA}.  }
    \item{\textbf{Difficulty:} \textit{minimal} number of search steps required among the correct traces from the search agent as an estimate of difficulty. If the number of search steps is greater than or equal to the target search step $S$, we consider the generated data as difficult enough. }
\end{itemize}
Algorithm \ref{alg:search_agent_verification} describes this verification process. 

Generating complex ($q$, $a$) pairs requiring multiple search steps is non-trivial as it involves interaction with the retrieval tool. As revealed by our evaluation in Table \ref{tab:intrinsic_results}, the data generator alone is only able to generate 18\% of the data that satisfy both the correctness and difficulty constraint, when tasked to generate questions that require 3 to 7 steps to solve. We visualize the per-step performance in Figure \ref{fig:performance_per_step}, which further reveals that the data generator fails more frequently when attempting to generate questions requiring more search steps.

\subsection{Generation with Execution Feedback}

Failure to generate a correct ($q$, $a$) pair requiring the target number of search steps reveals a discrepancy between the data generator's trajectory and search agent's trajectory. For instance, two steps planned by the data generator might be solved with a single search step by the search agent, as we later analyze in Section \ref{subsec:error_analysis}. Can we leverage \textit{both} trajectories to reconcile such discrepancy? Instead of merely leveraging search agent's execution results as a filter \citep{Shi2025PanguDA}, we propose to leverage the execution traces as \textit{feedback} for the data generator. 

Concretely, we feed both the data generator's traces and the search agent's traces, each containing the retrieved documents, back to the data generator. which is now tasked to output an updated (question, answer) pairs (Line \ref{alg:feedback_start}-\ref{alg:feedback_end} in Alg~\ref{alg:data_generation}). This process can be conducted in an iterative manner, alternating between generating a new question and collecting execution feedback from the search agent for the updated ($q$, $a$) pair. Finally, we filter out ($q$, $a$) pairs for which none of the execution trace from the search agent arrives at the same answer as the data generator's answer (pass@K=0).

\section{Experiments}
We conduct intrinsic evaluation of our methods by measuring the correctness and difficulty of the generated data in Section \ref{sec:intrinsic_eval}. We further conduct downstream evaluation in Section \ref{sec:extrinsic_eval} by training search agents on our generated data.

\subsection{Experiment setting}

\paragraph{Input corpus and retrieval setting.} We use the 2018 Wikipedia dump as our input corpus~\citep{karpukhin-etal-2020-dense}. Following \cite{jin2025search}, we use E5~\citep{Wang2022TextEB} as the retriever module. The number of returned passages for each search call is set to 3.

\paragraph{Data generation pipeline setting.} We use gemini-2.5-flash  as the LLM which acts as both the data generator and the search agent. The maximum number of search steps are set to 20 for both the data generator model and the search agent. We report results with input target $S$ set to 3 to 7 steps. 

\subsection{Intrinsic evaluation}\label{sec:intrinsic_eval}

\begin{table}
\begin{center}
\footnotesize
\begin{tabular}{@{}lrrrr@{}}
\toprule
\multirow{2}{*}{\textbf{System}} & \multicolumn{2}{c}{\textbf{Data Quality}} & \multicolumn{2}{c}{\textbf{Difficulty}} \\
&  {\% corr $\uparrow$} & {\% pass$\uparrow$} & {Avg@4$\downarrow$} & {\# search$\uparrow$} \\ 
 \midrule
 \multicolumn{5}{c}{\textit{\textbf{Baseline}}} \\
$A_{gen}$ w/o $S$ & 84 & - & 86.3 & 3.2 \\
$A_{gen}$ & 71 & 18 & 87.4 & 3.3 \\
 +1 resample & 77 & 27 & 84.5 & 3.8 \\ 
 +2 resample & 81 & 38 & 80.3 & 4.3 \\ 
 +3 resample & 84 & 47 & 80.1 & 4.8 \\ 
\midrule 
\multicolumn{5}{c}{\textit{\textbf{Ours: SAGE}}} \\
 +1 feedback & 77 & 31 & 83.2 & 4.1 \\ 
 +2 feedback & 83 & 42 & 80.4 & 4.6 \\ 
 +3 feedback & \textbf{87} & \textbf{50} & \textbf{79.5} & \textbf{4.9} \\ 
\bottomrule
\end{tabular} 
\end{center}
\caption{Evaluation of data quality. We report the portion of correct (\textbf{\% corr}) and portion of successful (\textbf{\% pass}: data that are correct and require at least $S$ steps to solve) generation out of all generated data. We measure difficulty of the correct portion of the generated data (pass@4=True) by Avg@4 and \# search steps needed. }
\label{tab:intrinsic_results}
\end{table}

\paragraph{Baselines.} We compare our method with (1) \textbf{Data generator without difficulty prompt: } we prompt the data generator to generate a complicated question which requires multiple steps to solve without pre-specifying a target search step $S$; (2) \textbf{Initial data generator model:} we present the result of the initial data generator model; (3) \textbf{Resampling}: instead of using execution feedback to update the (question, answer) pair, we re-sample another pair from $A_{gen}$ for samples that are incorrect or not difficult enough. This is similar to best-of-K sampling with the search agent's execution result as the verifier. Note that both the resampling baseline and \texttt{SAGE} can be conducted in multiple rounds. We report results for up to 3 rounds.

\paragraph{Evaluation.} We report various metric to measure the quality of the generated data:  \textbf{\% correct}: this reports the proportion of data out of all generated data that has pass@K=1, we set K as 4; \textbf{\% pass}: this measures the proportion of data that is both correct and require the search agent at least $S$ step to solve. Note that this metric is undefined for the baseline with no difficulty control. We further report two difficulty metrics for the questions that are correct: \textbf{Avg@4}: this measures the average performance of the search agent out of 4 traces. A lower \textbf{Avg@4} indicates a more difficult question. \textbf{Number of search steps:} the number of search steps required by the search agent to solve the question. We estimate this with the \textit{minimal} number out of the correct traces. For correctness, we use LLM-as-a-judge against the answer proposed by the data generator as the reference answer.\footnote{We use \texttt{gemini-2.0-flash} as the LLM and the exact prompt is included in Table \ref{fig:llm_as_judge_prompt}}

\begin{figure}
    \centering
    \includegraphics[width=0.7\textwidth]{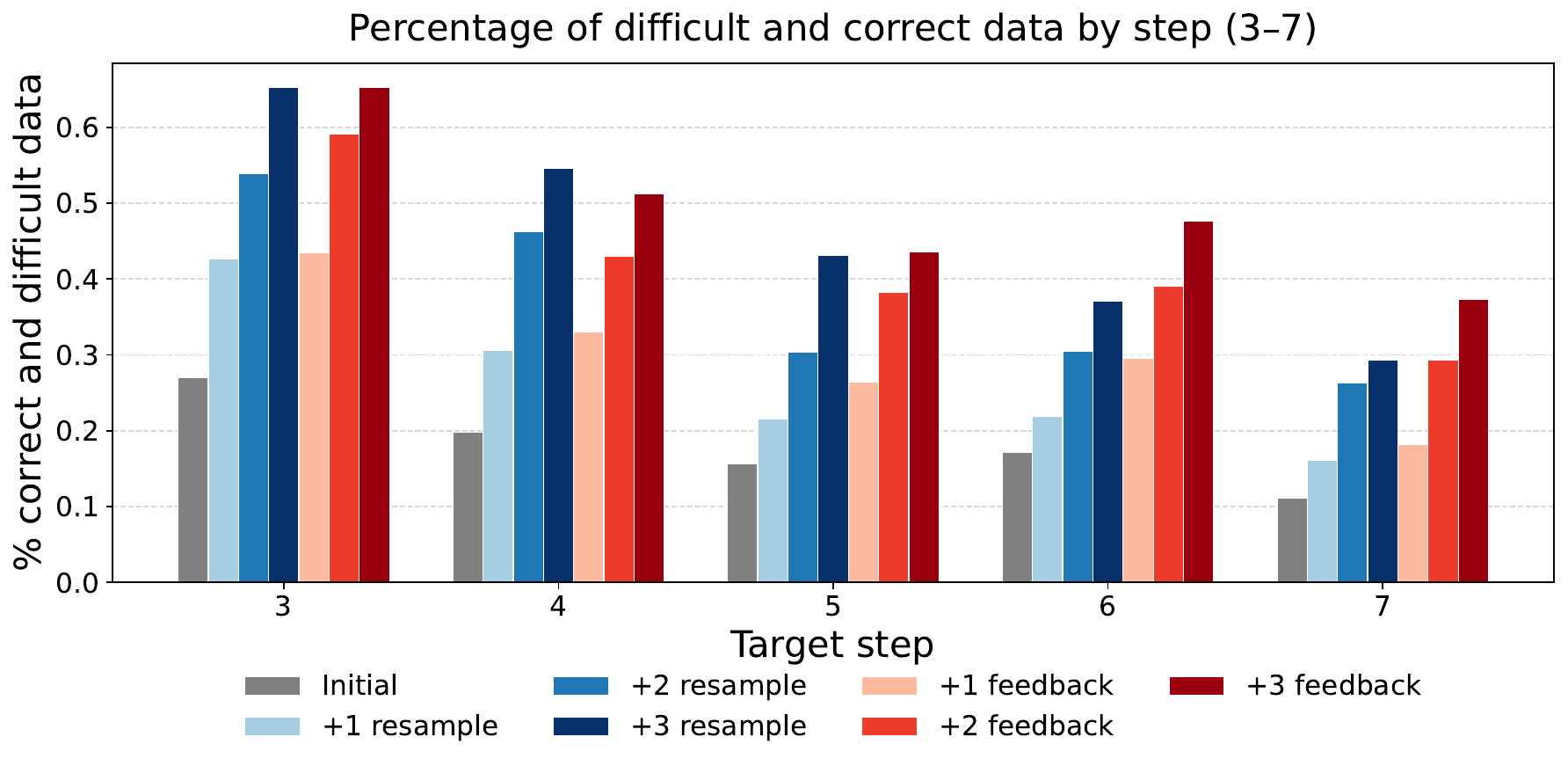}
    \caption{Percentage of correct and difficult data for target steps of 3 to 7 for various methods. Updating the question with execution feedback further improves the data quality compared to resampling, particularly with more target search steps.}
    \label{fig:performance_per_step}
\end{figure}

\paragraph{Results.} 
Table \ref{tab:intrinsic_results} summarizes the results. Comparing $A_{gen}$ with its variant without the target step $S$, we observe that explicitly including the target step in the prompt leads to a slight increase in the number of search steps required. However, $A_{gen}$ alone struggles to generate questions that require the specified number of steps, achieving only an 18\% pass rate. Resampling and incorporating execution feedback substantially improve the success rate and yield more challenging questions overall. Among the two strategies, execution feedback consistently outperforms resampling across different numbers of rounds. Figure \ref{fig:performance_per_step} further breaks down performance by target step, showing that feedback provides greater benefits than resampling as the target step increases, indicating its effectiveness for generating more complex questions. We thus only use samples generated with execution feedback in our downstream evaluation. We include example questions generated by \texttt{SAGE} in Table \ref{tab:example_question} in the Appendix.

\begin{table*}
\begin{center}
\footnotesize
\begin{tabular}{@{}ll|rrrrrr|rrr@{}}
\toprule

\multirow{2}{*}{\textbf{Training Data}} & \multirow{2}{*}{\textbf{Backbone Model}} & \multicolumn{6}{c|}{\textbf{In-domain}} & \multicolumn{2}{c}{\textbf{Out-of-domain}} \\
& & 3-hop & 4-hop & 5-hop & 6-hop & 7-hop & AVG & {Musique} & {FRAMES}   \\
\midrule
- & gemini-2.0-flash & 68.7 & 55.7 & 50.3 & 43.3 & 41.3 & 51.9 & 25.2 & 45.9 \\
- & gemini-2.5-flash & 80.0 & 67.0 & 57.3 & 48.7  & 37.3 & 58.1 &  28.0 & 50.0 \\ 
\midrule 
NQ + HotpotQA & QWEN-3B  & 25.3 & 12.0 & 15.3 & 11.7 & 15.0 & 15.9 & 11.4 & 13.3 \\ 
Musique & QWEN-3B & 37.3 & 20.0 & 18.3 & 19.0 & 17.1 & 22.4 & 19.4 & 21.5 \\ 
Ours & QWEN-3B & \textbf{42.3} & \textbf{26.7} & \textbf{25.3} & \textbf{25.0} & \textbf{23.3} & \textbf{28.5} & \textbf{19.9} & \textbf{23.8} \\
\midrule 
NQ + HotpotQA & QWEN-7B &45.0 & 26.3 & 25.7 & 24.7 & 23.6 & 29.1 &  18.9 & 26.2 \\
Musique & QWEN-7B & 48.3 & 28.7 & 24.7 & 25.0 & 21.2 & 29.6 & {21.6} & 25.0 \\
Ours & QWEN-7B & \textbf{55.7} &\textbf{38.0}&  \textbf{35.7} & \textbf{37.3} & \textbf{24.0} & \textbf{38.1} & \textbf{22.3} & \textbf{32.3} \\ 
\bottomrule
\end{tabular} 
\end{center}
\caption{Downstream evaluation: performance of search agents on complex question answering datasets. We report performance using LLM as a judge against reference answers. Each of the data split contains 300 sampled data points. All models search for up to 10 turns.
}
\label{tab:downstream_results}
\end{table*}

\subsection{Downstream evaluation}\label{sec:extrinsic_eval}

We conduct extrinsic evaluation of data generated by \texttt{SAGE} by using it to train search agents. We adopt the Search-R1~\citep{jin2025search} framework to train the agents and compare with agents trained with existing public data.

\paragraph{Baselines.} We consider models trained on publicly available question answering data: (1) \textbf{NQ + HotpotQA}: a combination of Natural Question~\citep{kwiatkowski-etal-2019-natural} and HotpotQA~\citep{yang-etal-2018-hotpotqa}, which consists of 150K data. This is the training data used by Search-R1~\citep{jin2025search}; (2) \textbf{Musique}~\citep{trivedi-etal-2022-musique}: consisting of multi-hop questions requiring two to four hops, the training split contains 20K data. We also compare with the performance of the prompting-based search agents using gemini-2.0-flash and gemini-2.5-flash as the agent.

\paragraph{Evaluation setting.} We evaluate on various downstream question answering datasets that are grounded in Wikipedia: (1) \textbf{In-domain}: we report performance for testing data generated with \texttt{SAGE}. We generate 300 test data for each target search step and report the per-step performance as well as the average performance. We also consider two out-of-domain datasets: (1) \textbf{Musique}: which contains 2 to 4-hop questions; we randomly sample 300 data for each hop and report the average performance. (2) \textbf{FRAMES}~\citep{krishna-etal-2025-fact}: which contains multi-hop questions created by human annotators. We report the results on a randomly sampled set of 300 questions. We evaluate the performance  with reference-based LLM-as-a-judge using \texttt{gemini-2.0-flash}, which outputs a binary judgment given the reference answer.

\paragraph{Data generation setting.} We generate data using gemini-2.5-flash as both the data generator and search agent. Similar to Section \ref{sec:intrinsic_eval}, we use E5 as the retriever system and the 2018 Wikipedia dump as the corpus. For all dataset generation settings, we filter out questions that require less than two search steps and generate 20K data for training, such that the training set size is the same as our baselines. We report results for data generated with two rounds of feedback and conduct ablation studies on using data generated with different number of rounds in Section \ref{subsec:ablation_on_round}.

\paragraph{Model and training setting.} We conduct experiment with two model variants: Qwen-2.5-3B-Instruct and Qwen-2.5-7B-Instruct~\citep{Yang2024Qwen25TR}. For \textbf{NQ+HotpotQA}, we directly inference with the Search-R1 checkpoints\footnote{\url{https://huggingface.co/PeterJinGo/SearchR1-nq_hotpotqa_train-qwen2.5-7b-it-em-ppo} and \\ \url{PeterJinGo/SearchR1-nq_hotpotqa_train-qwen2.5-3b-em-ppo}.}. For models trained with musique and our data, we train the models using reinforcement learning with the reward defined as reference-based LLM-as-a-judge using \texttt{gemini-2.0-flash}. All models are trained with PPO~\citep{schulman2017proximal}. Following prior work~\cite{jin2025search}, we apply loss masking to the retrieved information. We include the implementation details in Section \ref{app:implementation_details_search_agent} in the Appendix.

\paragraph{Retrieval settings.} For all training experiments, we use E5 as the retriever and the 2018 Wikipedia dump as the corpus, matching the set-up used for data generation. For inference, we use the same set-up except when evaluating on FRAMES, where we use the 2023 Wikipeida dump\footnote{\url{https://huggingface.co/datasets/wikimedia/wikipedia/viewer/20231101.en}} to align with its data construction process~\citep{krishna-etal-2025-fact}.

\paragraph{Results.} Table \ref{tab:downstream_results} reports downstream performance of models trained on different QA datasets. Across both QWEN-3B and QWEN-7B, training on our generated data yields consistent gains over NQ + HotpotQA and Musique on in-domain evaluations. For QWEN-3B, training on our data increases average accuracy from 15.9\% (NQ + HotpotQA) and 22.4\% (Musique) to 28.5\%, showing 27\% relative improvement.  For QWEN-7B, while training on Musique provides only a marginal improvement over NQ + HotpotQA (29.6\% vs. 29.1\%), training on our data further boosts average accuracy to 38.1\%, leading to 29\% relative improvement. Training on our data also improves performance on out-of-domain datasets. On FRAMES, training on our data enables relative improvement of 11\% for QWEN-3B and 23\% for QWEN-7B. Notably, on Musique, QWEN-7B trained on our data achieves higher accuracy (22.3\%) then directly on the in-domain Musique data (21.6\%).

\subsection{Evaluation with Google Search}\label{subsec:google_search}
While our method generates challenging ($q$, $a$) pairs using a fixed corpus (e.g. Wikipedia), we further ask whether search agents trained with retrieval over a fixed corpus can generalize to other retrieval tool. To this end, we evaluate agents trained with Wikipedia-based retrieval on benchmarks that require Google search.

\paragraph{Datasets and settings.} We evaluate on three benchmarks that require Google search: (1) \textbf{GAIA}~\citep{mialon2023gaiabenchmarkgeneralai}; (2) \textbf{Browsecomp}~\citep{Wei2025BrowseCompAS} and (3) \textbf{Humanity's Last Exam(HLE)-search}~\citep{phan2025humanitysexam}. For GAIA, we report results on the text-only subset with 103 questions. For Browsecomp, we evaluate on a randomly sampled subset of 200 questions. For HLE, we evaluate on a subset of questions that required search as classified by Gemini-1.5-pro, following prior work~\citep{han2025deepresearchertesttimediffusion}. 
During inference, we replace Wikipedia-based retrieval with Google Search using the Serper API \footnote{\url{https://serper.dev/}}, retrieving the top three snippets per search query. 

\paragraph{Results.} Table \ref{tab:google_search_results} reports the results. We observe a promising trend indicating that training search agents with retrieval over a fixed corpus (e.g. Wikipedia) enables effective transfer when using a different retrieval tool (e.g. Google Search). Training on data generated by our pipeline leads to substantial improvements on GAIA compared to training on existing large-scale training data, for both 3B and 7B models (36\% and 50\% relative improvements compared to the strongest baseline respectively). We also observe improvement on Browsecomp, which consists of multi-step search questions, for the QWEN-7B model. QWEN-3B achieves similarly poor performance (1.0\% accuracy) across the three training datasets, likely because the questions are too challenging for a 3B model. On HLE, gains are more modest for both models, likely due to the substantial domain shift toward highly specialized scientific questions.

\begin{table}
\begin{center}
\footnotesize
\begin{tabular}{@{}lrrr@{}}
\toprule
\textbf{Training data}   &{\textbf{GAIA}} & \textbf{Browsecomp} & {\textbf{HLE-Search}} \\
\midrule
\multicolumn{4}{c}{\textit{\textbf{QWEN-3B}}} \\
 NQ + HotpotQA   & 12.5 & 1.0 & 5.0 \\ 
 Musique  & 13.5 & 1.0 & 4.0 \\ 
 Ours & \textbf{18.8} & 1.0 & \textbf{5.5}  \\
\midrule 
\multicolumn{4}{c}{\textit{\textbf{QWEN-7B}}} \\
 NQ + HotpotQA   & 14.6 & 1.6 & 4.5 \\ 
 Musique  & 15.6 & 2.1 & \textbf{8.0} \\ 
  Ours  & \textbf{24.0} & \textbf{2.6} & 7.0 \\
\bottomrule
\end{tabular} 
\end{center}
\caption{Results on deep search benchmarks using Google Search, with a maximum of 10 queries for GAIA and 20 for other benchmarks.}
\label{tab:google_search_results}
\end{table}

\section{Analysis}

\subsection{Ablation on feedback rounds}\label{subsec:ablation_on_round}
In Section \ref{sec:extrinsic_eval}, we report downstream evaluation on training with data generated with \texttt{SAGE} with two rounds of feedback. How does the number of rounds impact downstream performance? We conduct an ablation study on training QWEN-7B model on data generated with 0-3 rounds of feedback. For all settings, we keep the training data size the same (20K) and filter out questions that require less than two search steps.

Results are reported in Table \ref{tab:ablation_results}. We report downstream performance on the in-domain testing set, averaged across questions requiring 2-7 search steps, as well as on Musique and FRAMES. We additionally report difficulty of the generated data, measured by \textbf{Avg@4}. We see improvement for both in-domain and out-of-domain datasets when increasing the number of feedback rounds from 0 to 2, confirming the effectiveness of incorporating execution feedback. Note that zero feedback round is equivalent to using the initial data generator only. We do not see further improvement when increasing the number of feedback to three rounds, despite that the data generated with three rounds of feedback being more difficult compared to those with two rounds. This suggests that increasing data difficulty alone is insufficient and highlights the need for more principled data curation strategies that balance difficulty and learnability.

\begin{table}
\begin{center}
\footnotesize
\begin{tabular}{@{}lrrrr@{}}
\toprule
\multirow{2}{*}{\textbf{Round}} & \multicolumn{3}{c}{\textbf{Downstream Performance}} &  \textbf{Difficulty}\\ 
& {\textbf{In-domain}} & \textbf{Musique} & \textbf{FRAMES} & \textbf{Avg@4$\downarrow$} \\
\midrule
0  & 33.6 & 18.7 & 29.0 & 86.3 \\ 
1 & 33.6 & 19.5 &  29.3 & 83.2\\ 
2 & \textbf{38.1} & \textbf{22.3} & \textbf{32.3} & 80.4 \\ 
3 & 34.1 & 20.9 & 28.1 & \textbf{79.5}  \\
\bottomrule
\end{tabular} 
\end{center}
\vspace{-0.3cm}
\caption{Ablation on training QWEN-7B on data generated with different numbers of feedback rounds.}
\label{tab:ablation_results}
\end{table}

\subsection{Reasoning strategy analysis}
\begin{figure}
    \centering
    \includegraphics[width=0.7\textwidth]{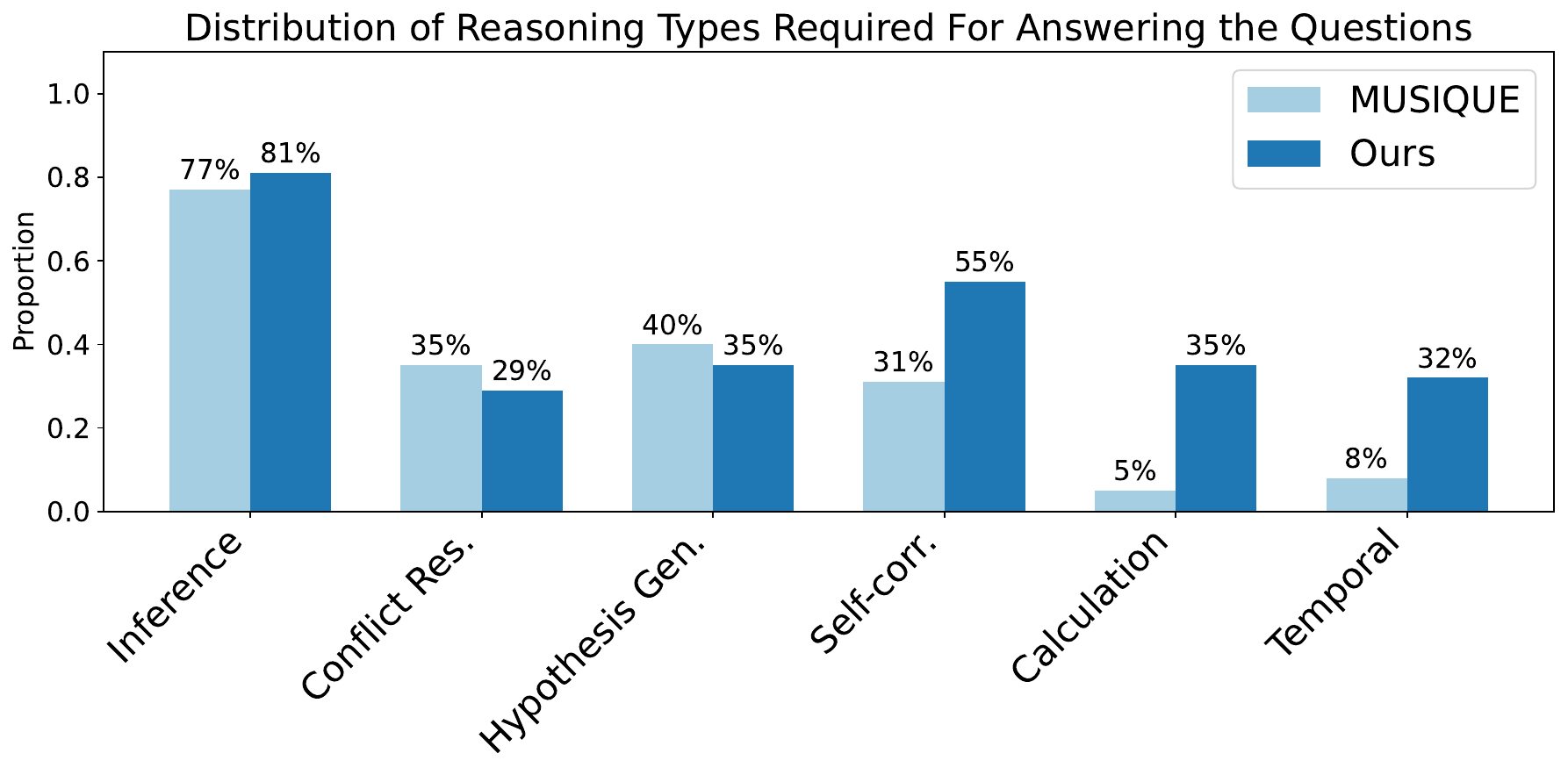}
    \vspace{-0.3cm}
    \caption{Distribution of reasoning types required to answer questions in Musique and our generated dataset. Our data spans a broad set of reasoning types, resulting in a more balanced distribution compared to Musique.}
    \label{fig:reasoning_strategies}
\vspace{-0.3cm}
\end{figure}

Solving deep search questions require the agent to be able to issue search queries as well as \textit{reason} about the retrieved documents. We conduct an analysis focusing on the reasoning strategies required to solve questions generated by our pipeline.

\paragraph{Setting.} We collect search trajectory by prompting gemini-2.5-flash as the search agent. We sample 100 trajectories that lead to the correct answer for our generated data and Musique respectively. We then prompt gemini-2.5-flash to first (1) identify the types of reasoning strategies present in these search trajectories and (2) for each trajectory, label the presence of the reasoning strategies for each step. We include the prompt, definition for each strategy, as well as implementation details of the analysis in Section \ref{app:analysis_details} in the Appendix.

\paragraph{Results.} Figure \ref{fig:reasoning_strategies} shows the distribution of reasoning types required to solve questions in our dataset and Musique, note that a single question typically requires multiple strategies. \texttt{SAGE} produces questions that span a broader range of reasoning types, including conflict resolution, hypothesis generation, self-correction, calculation, and temporal reasoning. In particular, calculation and temporal reasoning are rare in Musique (5\% and 8\%, respectively) but appear more frequently in our data (35\% and 32\%), resulting in a more balanced distribution across reasoning categories.

\subsection{Analyzing data generator's error}\label{subsec:error_analysis}
\begin{table}
\begin{center}
\small
\begin{tabular}{@{}lr@{}}
\toprule
\textbf{Category} & \textbf{Percentage} \\ \midrule
\multicolumn{2}{c}{\textit{Easy data}} \\
Superficial complexity & 13\% \\ 
Multi-query collapse & 21\% \\  
Overly specific question & 31\% \\ 
Information co-location & 35\% \\ \midrule
\multicolumn{2}{c}{\textit{Incorrect data}} \\
Ambiguous question &  7\% \\ 
Data generator error & 19\% \\ 
Search agent error & 20\% \\ 
Search agent retrieval failure &  54\% \\ 
\bottomrule
\end{tabular} 
\end{center}
\caption{Error categorization for ($q$, $a$) pair generated by the initial data generator.}
\label{tab:error_pattern_categorization}
\vspace{-0.3cm}
\end{table}

To better understand the main causes of the data generator's failure to produce correct and appropriately difficult ($q$, $a$) pairs, we analyze the discrepancies between the data generator's trajectories and the execution traces of the search agent.
Specifically, we prompt gemini-2.5-flash to compare these two traces and categorize the error into error patterns. We identify four common failure patterns for the two error type (easy data and incorrect data). Results are reported in Table \ref{tab:error_pattern_categorization} and the prompt used for the analysis is in Section \ref{app:analysis_details} in the appendix.

For easy data, we observe that most errors stem from misalignment between the data generator's intended difficulty and the actual difficulty required by the search agent to solve the question, which is often influenced by the external environment. For instance, 35\% of the generated question exhibit \textbf{information co-location}, where multiple pieces of information required to answer the question occurs in the same document in the corpus; 21\% of the generated data exhibits \textbf{multi-query collapse}, in which information from \textit{different documents} can be retrieved by the retrieval tool using a single query. Both phenomena reduce the number of search steps required to answer the question. This analysis motivates our method as such discrepancy is only discoverable via execution feedback.

For incorrect data, the misalignment between the search agent's answer and the data generator's answer most commonly arises from failure on the search agent, including retrieval failure or reasoning errors. 
While we do not distinguish these data from truly incorrect data and filter them out for training, future work could explore verifying the correctness of ($q$, $a$) pairs unsolved by the search agent. Around 20\% of the incorrect data is attributable to errors made by the data generator, such as hallucinating ungrounded steps in the question generation process. The remaining 7\% of data involves ambiguous question which leads to different answer found by the search agent.

\section{Related Work}\label{sec:related_work}

\paragraph{Deep Search.} Early retrieval-augmented generation (RAG) systems typically perform a single retrieval step before generating an answer \citep{NEURIPS2020_6b493230, Gao2023RetrievalAugmentedGF}. Recent agentic RAG approaches aim to support deep search, in which models iteratively interleave retrieval and reasoning to solve questions requiring multiple search steps. 
\citet{trivedi-etal-2023-interleaving} is among the earlier work that explore a prompting-only strategy that guides model to alternate between search and reasoning until reaching an answer. Search-o1 \citep{search-o1} furthre improves deep search by incorporating in-document reasoning to filter out irrelevant retrieved information. Search-R1 \citep{jin2025search} proposes training search agents following the ReACT framework~\citep{yao2023react} using outcome-based reward. Subsequent works~\citep{pangu-deepdiver, webdancer} all adopt RL training paradigm as high-quality SFT trajectories for deep search agent is difficult to collect at scale. We also adopt the Search-R1 training framework for our downstream evaluation.

\paragraph{Synthetic data generation for deep search.} 
We review several concurrent works on synthetic data generation for deep search. WebDancer~\citep{webdancer} adopt reverse QA generation and difficulty filtering, using either web pages or Wikipedia, while  WebShaper~\citep{webshaper} rely on structured signals such as knowledge graphs or explicit entity relations to construct complex questions. WebSailor~\citep{li2025websailornavigatingsuperhumanreasoning}, WebPuzzle~\citep{pangu-deepdiver} and WebExplorer~\citep{liu2025webexplorerexploreevolvetraining} generate challenging questions by progressively obscuring information around seed entities. In contrast, our approach takes in a random passage and leverages strong LLM-based data generator and search agent to flexibly synthesize challenging ($q$, $a$) pairs without relying on high-quality knowledge graphs or manually imposed structure.

We further note that many concurrent works~\citep{webdancer, pangu-deepdiver, webshaper, liu2025webexplorerexploreevolvetraining, li2025websailornavigatingsuperhumanreasoning} depend on commercial retrieval APIs (e.g., Google Search) during data generation. This substantially increases the cost for both data generation and model training due to the large number of intermediate search queries required.  Our pipeline is instead grounded in a fixed corpus, while experiments in Section~\ref{subsec:google_search} demonstrate promising transfer to using Google Search at inference time only. We summarize a comparison between our work and the concurrent works in Table \ref{tab:concurrent_work_comparison} in the appendix. Finally, existing baseline training datasets~\citep{kwiatkowski-etal-2019-natural, yang-etal-2018-hotpotqa, trivedi-etal-2022-musique} are constructed either through human annotation, automatic pipeline leveraging Wikipedia structure or a combination of both. To our best knowledge, these are the only other publicly available training data for search agents grounded in Wikipedia.

\section{Conclusion}
We introduce an agentic pipeline, \texttt{SAGE}, to automatically generate deep search data for a given corpus. Our framework consists of a data generator agent that generates complicated question as well as answers, and a search agent that provides execution feedback by attempting to solve the question. We conduct comprehensive evaluation on both intrinsic data quality and downstream evaluation, training search agents with our generated data. Evaluation on in-domain and out-of-domain evaluation set demonstrates the effectiveness of our framework.

\section*{Limitations}

\paragraph{Proposed method.} Our dual-agent framework currently relies on a fixed search agent to provide execution feedback for the data generator agent to evolve the generated data. Future work can explore \textit{co-evolving} both agents (e.g., through iterative training), which could further enhance the quality of the generated data and hence the capability of the search agent. We adopt pass@K=1 as the correctness criterion for the generated data, which serves as a practical approximation but may still admit hallucinated or incorrect content. Future work could investigate more robust verification methods for identifying correct ($q$, $a$) pairs that the existing search agent cannot solve (pass@K=0). Our work also only focuses on generating high-quality $(q, a)$ pairs for RL training. Exploring methods to construct high quality supervised fine-tuning trajectories to train search agents can be a promising direction.

\paragraph{Experimental setting.} While we evaluate our method using both intrinsic evaluation and downstream evaluation, our experiments do not explore alternative reinforcement learning algorithms such as GRPO~\citep{Shao2024DeepSeekMathPT}, nor model scales beyond 7B parameters. We only experiment with generating data from a single general domain corpus, Wikipedia. Future work can explore generating deep search data for other domain-specific corpus, such as legal or scientific domains.

\section*{Acknowledgments}
We thank Tengxiao Liu, Siru Ouyang, Yihe Deng, Rui Meng and other members from Google Cloud AI Research for helpful discussions and feedbacks throughout the project.

\bibliographystyle{abbrvnat}
\nobibliography*
\bibliography{custom}

@inproceedings{
zhou2024webarena,
title={WebArena: A Realistic Web Environment for Building Autonomous Agents},
author={Shuyan Zhou and Frank F. Xu and Hao Zhu and Xuhui Zhou and Robert Lo and Abishek Sridhar and Xianyi Cheng and Tianyue Ou and Yonatan Bisk and Daniel Fried and Uri Alon and Graham Neubig},
booktitle={The Twelfth International Conference on Learning Representations},
year={2024},
url={https://openreview.net/forum?id=oKn9c6ytLx}
}

@inproceedings{
jimenez2024swebench,
title={{SWE}-bench: Can Language Models Resolve Real-world Github Issues?},
author={Carlos E Jimenez and John Yang and Alexander Wettig and Shunyu Yao and Kexin Pei and Ofir Press and Karthik R Narasimhan},
booktitle={The Twelfth International Conference on Learning Representations},
year={2024},
url={https://openreview.net/forum?id=VTF8yNQM66}
}

@inproceedings{
jin2025search,
title={Search-R1: Training {LLM}s to Reason and Leverage Search Engines with Reinforcement Learning},
author={Bowen Jin and Hansi Zeng and Zhenrui Yue and Jinsung Yoon and Sercan O Arik and Dong Wang and Hamed Zamani and Jiawei Han},
booktitle={Second Conference on Language Modeling},
year={2025},
url={https://openreview.net/forum?id=Rwhi91ideu}
}

@inproceedings{
yao2023react,
title={ReAct: Synergizing Reasoning and Acting in Language Models},
author={Shunyu Yao and Jeffrey Zhao and Dian Yu and Nan Du and Izhak Shafran and Karthik R Narasimhan and Yuan Cao},
booktitle={The Eleventh International Conference on Learning Representations },
year={2023},
url={https://openreview.net/forum?id=WE_vluYUL-X}
}

@inproceedings{
asai2024selfrag,
title={Self-{RAG}: Learning to Retrieve, Generate, and Critique through Self-Reflection},
author={Akari Asai and Zeqiu Wu and Yizhong Wang and Avirup Sil and Hannaneh Hajishirzi},
booktitle={The Twelfth International Conference on Learning Representations},
year={2024},
url={https://openreview.net/forum?id=hSyW5go0v8}
}

@article{Nakano2021WebGPTBQ,
  title={WebGPT: Browser-assisted question-answering with human feedback},
  author={Reiichiro Nakano and Jacob Hilton and Suchir Balaji and Jeff Wu and Ouyang Long and Christina Kim and Christopher Hesse and Shantanu Jain and Vineet Kosaraju and William Saunders and Xu Jiang and Karl Cobbe and Tyna Eloundou and Gretchen Krueger and Kevin Button and Matthew Knight and Benjamin Chess and John Schulman},
  journal={ArXiv},
  year={2021},
  volume={abs/2112.09332},
  url={https://api.semanticscholar.org/CorpusID:245329531}
}

@article{kwiatkowski-etal-2019-natural,
    title = "Natural Questions: A Benchmark for Question Answering Research",
    author = "Kwiatkowski, Tom  and
      Palomaki, Jennimaria  and
      Redfield, Olivia  and
      Collins, Michael  and
      Parikh, Ankur  and
      Alberti, Chris  and
      Epstein, Danielle  and
      Polosukhin, Illia  and
      Devlin, Jacob  and
      Lee, Kenton  and
      Toutanova, Kristina  and
      Jones, Llion  and
      Kelcey, Matthew  and
      Chang, Ming-Wei  and
      Dai, Andrew M.  and
      Uszkoreit, Jakob  and
      Le, Quoc  and
      Petrov, Slav",
    editor = "Lee, Lillian  and
      Johnson, Mark  and
      Roark, Brian  and
      Nenkova, Ani",
    journal = "Transactions of the Association for Computational Linguistics",
    volume = "7",
    year = "2019",
    address = "Cambridge, MA",
    publisher = "MIT Press",
    url = "https://aclanthology.org/Q19-1026/",
    doi = "10.1162/tacl_a_00276",
    pages = "452--466",
}

@inproceedings{yang-etal-2018-hotpotqa,
    title = "{H}otpot{QA}: A Dataset for Diverse, Explainable Multi-hop Question Answering",
    author = "Yang, Zhilin  and
      Qi, Peng  and
      Zhang, Saizheng  and
      Bengio, Yoshua  and
      Cohen, William  and
      Salakhutdinov, Ruslan  and
      Manning, Christopher D.",
    editor = "Riloff, Ellen  and
      Chiang, David  and
      Hockenmaier, Julia  and
      Tsujii, Jun{'}ichi",
    booktitle = "Proceedings of the 2018 Conference on Empirical Methods in Natural Language Processing",
    month = oct # "-" # nov,
    year = "2018",
    address = "Brussels, Belgium",
    publisher = "Association for Computational Linguistics",
    url = "https://aclanthology.org/D18-1259/",
    doi = "10.18653/v1/D18-1259",
    pages = "2369--2380",
}

@article{trivedi-etal-2022-musique,
    title = "{M}u{S}i{Q}ue: Multihop Questions via Single-hop Question Composition",
    author = "Trivedi, Harsh  and
      Balasubramanian, Niranjan  and
      Khot, Tushar  and
      Sabharwal, Ashish",
    editor = "Roark, Brian  and
      Nenkova, Ani",
    journal = "Transactions of the Association for Computational Linguistics",
    volume = "10",
    year = "2022",
    address = "Cambridge, MA",
    publisher = "MIT Press",
    url = "https://aclanthology.org/2022.tacl-1.31/",
    doi = "10.1162/tacl_a_00475",
    pages = "539--554",
}

@article{Wei2025BrowseCompAS,
  title={BrowseComp: A Simple Yet Challenging Benchmark for Browsing Agents},
  author={Jason Wei and Zhiqing Sun and Spencer Papay and Scott McKinney and Jeffrey Han and Isabella Fulford and Hyung Won Chung and Alexandre Passos and William Fedus and Amelia Glaese},
  journal={ArXiv},
  year={2025},
  volume={abs/2504.12516},
  url={https://api.semanticscholar.org/CorpusID:277857238}
}

@article{schulman2017proximal,
  title={Proximal policy optimization algorithms},
  author={Schulman, John and Wolski, Filip and Dhariwal, Prafulla and Radford, Alec and Klimov, Oleg},
  journal={arXiv preprint arXiv:1707.06347},
  year={2017}
}

@inproceedings{trivedi-etal-2023-interleaving,
    title = "Interleaving Retrieval with Chain-of-Thought Reasoning for Knowledge-Intensive Multi-Step Questions",
    author = "Trivedi, Harsh  and
      Balasubramanian, Niranjan  and
      Khot, Tushar  and
      Sabharwal, Ashish",
    editor = "Rogers, Anna  and
      Boyd-Graber, Jordan  and
      Okazaki, Naoaki",
    booktitle = "Proceedings of the 61st Annual Meeting of the Association for Computational Linguistics (Volume 1: Long Papers)",
    month = jul,
    year = "2023",
    address = "Toronto, Canada",
    publisher = "Association for Computational Linguistics",
    url = "https://aclanthology.org/2023.acl-long.557/",
    doi = "10.18653/v1/2023.acl-long.557",
    pages = "10014--10037",
}

@article{Wang2022TextEB,
  title={Text Embeddings by Weakly-Supervised Contrastive Pre-training},
  author={Liang Wang and Nan Yang and Xiaolong Huang and Binxing Jiao and Linjun Yang and Daxin Jiang and Rangan Majumder and Furu Wei},
  journal={ArXiv},
  year={2022},
  volume={abs/2212.03533},
  url={https://api.semanticscholar.org/CorpusID:254366618}
}

@article{Yang2024Qwen25TR,
  title={Qwen2.5 Technical Report},
  author={Qwen An Yang and Baosong Yang and Beichen Zhang and Binyuan Hui and Bo Zheng and Bowen Yu and Chengyuan Li and Dayiheng Liu and Fei Huang and Guanting Dong and Haoran Wei and Huan Lin and Jian Yang and Jianhong Tu and Jianwei Zhang and Jianxin Yang and Jiaxin Yang and Jingren Zhou and Junyang Lin and Kai Dang and Keming Lu and Keqin Bao and Kexin Yang and Le Yu and Mei Li and Mingfeng Xue and Pei Zhang and Qin Zhu and Rui Men and Runji Lin and Tianhao Li and Tingyu Xia and Xingzhang Ren and Xuancheng Ren and Yang Fan and Yang Su and Yi-Chao Zhang and Yunyang Wan and Yuqi Liu and Zeyu Cui and Zhenru Zhang and Zihan Qiu and Shanghaoran Quan and Zekun Wang},
  journal={ArXiv},
  year={2024},
  volume={abs/2412.15115},
  url={https://api.semanticscholar.org/CorpusID:274859421}
}

@article{Gao2023RetrievalAugmentedGF,
  title={Retrieval-Augmented Generation for Large Language Models: A Survey},
  author={Yunfan Gao and Yun Xiong and Xinyu Gao and Kangxiang Jia and Jinliu Pan and Yuxi Bi and Yi Dai and Jiawei Sun and Qianyu Guo and Meng Wang and Haofen Wang},
  journal={ArXiv},
  year={2023},
  volume={abs/2312.10997},
  url={https://api.semanticscholar.org/CorpusID:266359151}
}

@inproceedings{NEURIPS2020_6b493230,
 author = {Lewis, Patrick and Perez, Ethan and Piktus, Aleksandra and Petroni, Fabio and Karpukhin, Vladimir and Goyal, Naman and K\"{u}ttler, Heinrich and Lewis, Mike and Yih, Wen-tau and Rockt\"{a}schel, Tim and Riedel, Sebastian and Kiela, Douwe},
 booktitle = {Advances in Neural Information Processing Systems},
 editor = {H. Larochelle and M. Ranzato and R. Hadsell and M.F. Balcan and H. Lin},
 pages = {9459--9474},
 publisher = {Curran Associates, Inc.},
 title = {Retrieval-Augmented Generation for Knowledge-Intensive NLP Tasks},
 url = {https://proceedings.neurips.cc/paper_files/paper/2020/file/6b493230205f780e1bc26945df7481e5-Paper.pdf},
 volume = {33},
 year = {2020}
}

@article{search-o1,
  author       = {Xiaoxi Li and
                  Guanting Dong and
                  Jiajie Jin and
                  Yuyao Zhang and
                  Yujia Zhou and
                  Yutao Zhu and
                  Peitian Zhang and
                  Zhicheng Dou},
  title        = {Search-o1: Agentic Search-Enhanced Large Reasoning Models},
  journal      = {CoRR},
  volume       = {abs/2501.05366},
  year         = {2025},
  url          = {https://doi.org/10.48550/arXiv.2501.05366},
  doi          = {10.48550/ARXIV.2501.05366},
  eprinttype    = {arXiv},
  eprint       = {2501.05366},
  bibsource    = {dblp computer science bibliography, https://dblp.org}
}

@article{pangu-deepdiver,
  title={Pangu DeepDiver: Adaptive Search Intensity Scaling via Open-Web Reinforcement Learning},
  author={Wenxuan Shi and Haochen Tan and Chuqiao Kuang and Xiaoguang Li and Xiaozhe Ren and Chen Zhang and Hanting Chen and Yasheng Wang and Lifeng Shang and Fisher Yu and Yunhe Wang},
  journal={arXiv preprint arXiv:2505.24332},
  year={2025}
}

@article{webdancer,
title = {WebDancer: Towards Autonomous Information Seeking Agency},
author = {Wu, Jialong and Li, Baixuan and Fang, Runnan and Yin, Wenbiao and Zhang, Liwen and Tao, Zhengwei and Zhang, Dingchu and Xi, Zekun and Jiang, Yong and Xie, Pengjun and Huang, Fei and Zhou, Jingren},
year = {2025},
month = {05},
pages = {},
doi = {10.48550/arXiv.2505.22648}
}

@article{webshaper,
title = {WebShaper: Agentically Data Synthesizing via Information-Seeking Formalization},
author = {Tao, Zhengwei and Wu, Jialong and Yin, Wenbiao and Zhang, Junkai and Li, Baixuan and Shen, Haiyang and Li, Kuan and Zhang, Liwen and Wang, Xinyu and Jiang, Yong and Xie, Pengjun and Huang, Fei and Zhou, Jingren},
year = {2025},
month = {07},
pages = {},
doi = {10.48550/arXiv.2507.15061}
}

@article{Li2025Searcho1AS,
  title={Search-o1: Agentic Search-Enhanced Large Reasoning Models},
  author={Xiaoxi Li and Guanting Dong and Jiajie Jin and Yuyao Zhang and Yujia Zhou and Yutao Zhu and Peitian Zhang and Zhicheng Dou},
  journal={ArXiv},
  year={2025},
  volume={abs/2501.05366},
  url={https://api.semanticscholar.org/CorpusID:275405676}
}

@article{webmall,
  title={WebMall -- A Multi-Shop Benchmark for Evaluating Web Agents},
  author={Ralph Peeters and Aaron Steiner and Luca Schwarz and Julian Yuya Caspary and Christian Bizer},
  journal={ArXiv},
  year={2025},
  volume={abs/2508.13024},
  url={https://arxiv.org/abs/2508.13024}
}

@article{surveycodeagent,
  title={A Survey on Code Generation with LLM-based Agents},
  author={Yihong Dong and Xue Jiang and Jiaru Qian and Tian Wang and Kechi Zhang and Zhi Jin and Ge Li},
  journal={ArXiv},
  year={2025},
  volume={abs/2508.00083},
  url={https://arxiv.org/abs/2508.00083}
}

@InProceedings{JoshiTriviaQA2017,
     author = {Joshi, Mandar and Choi, Eunsol and Weld, Daniel S. and Zettlemoyer, Luke},
     title = {TriviaQA: A Large Scale Distantly Supervised Challenge Dataset for Reading Comprehension},
     booktitle = {Proceedings of the 55th Annual Meeting of the Association for Computational Linguistics},
     month = {July},
     year = {2017},
     address = {Vancouver, Canada},
     publisher = {Association for Computational Linguistics},
}

@article{han2024ragqaarena,
  title={RAG-QA Arena: Evaluating Domain Robustness for Long-form Retrieval Augmented Question Answering},
  author={Rujun Han and Yuhao Zhang and Peng Qi and Yumo Xu and Jenyuan Wang and Lan Liu and William Yang Wang and Bonan Min and Vittorio Castelli},
  year={2024},
  journal={arXiv preprint arXiv:2407.13998},
  url={https://arxiv.org/abs/2407.13998}
}

@article{searchqa,
  title={SearchQA: A New Q\&A Dataset Augmented with Context from a Search Engine},
  author={Matthew Dunn and Levent Sagun and Mike Higgins and V. Ugur Guney and Volkan Cirik and Kyunghyun Cho},
  year={2017},
  journal={arXiv preprint arXiv:1704.05179},
  url={https://arxiv.org/abs/1704.05179}
}

@inproceedings{krishna-etal-2025-fact,
    title = "Fact, Fetch, and Reason: A Unified Evaluation of Retrieval-Augmented Generation",
    author = "Krishna, Satyapriya  and
      Krishna, Kalpesh  and
      Mohananey, Anhad  and
      Schwarcz, Steven  and
      Stambler, Adam  and
      Upadhyay, Shyam  and
      Faruqui, Manaal",
    editor = "Chiruzzo, Luis  and
      Ritter, Alan  and
      Wang, Lu",
    booktitle = "Proceedings of the 2025 Conference of the Nations of the Americas Chapter of the Association for Computational Linguistics: Human Language Technologies (Volume 1: Long Papers)",
    month = apr,
    year = "2025",
    address = "Albuquerque, New Mexico",
    publisher = "Association for Computational Linguistics",
    url = "https://aclanthology.org/2025.naacl-long.243/",
    doi = "10.18653/v1/2025.naacl-long.243",
    pages = "4745--4759",
    ISBN = "979-8-89176-189-6",
}

@article{Schulman2015HighDimensionalCC,
  title={High-Dimensional Continuous Control Using Generalized Advantage Estimation},
  author={John Schulman and Philipp Moritz and Sergey Levine and Michael I. Jordan and P. Abbeel},
  journal={CoRR},
  year={2015},
  volume={abs/1506.02438},
  url={https://api.semanticscholar.org/CorpusID:3075448}
}

@inproceedings{karpukhin-etal-2020-dense,
    title = "Dense Passage Retrieval for Open-Domain Question Answering",
    author = "Karpukhin, Vladimir  and
      Oguz, Barlas  and
      Min, Sewon  and
      Lewis, Patrick  and
      Wu, Ledell  and
      Edunov, Sergey  and
      Chen, Danqi  and
      Yih, Wen-tau",
    editor = "Webber, Bonnie  and
      Cohn, Trevor  and
      He, Yulan  and
      Liu, Yang",
    booktitle = "Proceedings of the 2020 Conference on Empirical Methods in Natural Language Processing (EMNLP)",
    month = nov,
    year = "2020",
    address = "Online",
    publisher = "Association for Computational Linguistics",
    url = "https://aclanthology.org/2020.emnlp-main.550/",
    doi = "10.18653/v1/2020.emnlp-main.550",
    pages = "6769--6781",
}

@article{Shao2024DeepSeekMathPT,
  title={DeepSeekMath: Pushing the Limits of Mathematical Reasoning in Open Language Models},
  author={Zhihong Shao and Peiyi Wang and Qihao Zhu and Runxin Xu and Jun-Mei Song and Mingchuan Zhang and Y. K. Li and Yu Wu and Daya Guo},
  journal={ArXiv},
  year={2024},
  volume={abs/2402.03300},
  url={https://api.semanticscholar.org/CorpusID:267412607}
}

@article{Chen2025BrowseCompPlusAM,
  title={BrowseComp-Plus: A More Fair and Transparent Evaluation Benchmark of Deep-Research Agent},
  author={Zijian Chen and Xueguang Ma and Shengyao Zhuang and Ping Nie and Kai Zou and Andrew Liu and Joshua Green and Kshama Patel and Ruoxi Meng and Mingyi Su and Sahel Sharifymoghaddam and Yanxi Li and Haoran Hong and Xinyu Shi and Xuye Liu and Nandan Thakur and Crystina Zhang and Luyu Gao and Wenhu Chen and Jimmy Lin},
  journal={ArXiv},
  year={2025},
  volume={abs/2508.06600},
  url={https://api.semanticscholar.org/CorpusID:280565737}
}

@article{Shi2025PanguDA,
  title={Pangu DeepDiver: Adaptive Search Intensity Scaling via Open-Web Reinforcement Learning},
  author={Wenxuan Shi and Haochen Tan and Chuqiao Kuang and Xiaoguang Li and Xiaozhe Ren and Chen Zhang and Hanting Chen and Yasheng Wang and Lifeng Shang and Fisher Yu and Yunhe Wang},
  journal={ArXiv},
  year={2025},
  volume={abs/2505.24332},
  url={https://api.semanticscholar.org/CorpusID:279057242}
}

@misc{phan2025humanitysexam,
      title={Humanity's Last Exam}, 
      author={Long Phan and Alice Gatti and Ziwen Han and Nathaniel Li and Josephina Hu and Hugh Zhang and Chen Bo Calvin Zhang and Mohamed Shaaban and John Ling and Sean Shi and Michael Choi and Anish Agrawal and Arnav Chopra and Adam Khoja and Ryan Kim and Richard Ren and Jason Hausenloy and Oliver Zhang and Mantas Mazeika and Dmitry Dodonov and Tung Nguyen and Jaeho Lee and Daron Anderson and Mikhail Doroshenko and Alun Cennyth Stokes and Mobeen Mahmood and Oleksandr Pokutnyi and Oleg Iskra and Jessica P. Wang and John-Clark Levin and Mstyslav Kazakov and Fiona Feng and Steven Y. Feng and Haoran Zhao and Michael Yu and Varun Gangal and Chelsea Zou and Zihan Wang and Serguei Popov and Robert Gerbicz and Geoff Galgon and Johannes Schmitt and Will Yeadon and Yongki Lee and Scott Sauers and Alvaro Sanchez and Fabian Giska and Marc Roth and Søren Riis and Saiteja Utpala and Noah Burns and Gashaw M. Goshu and Mohinder Maheshbhai Naiya and Chidozie Agu and Zachary Giboney and Antrell Cheatom and Francesco Fournier-Facio and Sarah-Jane Crowson and Lennart Finke and Zerui Cheng and Jennifer Zampese and Ryan G. Hoerr and Mark Nandor and Hyunwoo Park and Tim Gehrunger and Jiaqi Cai and Ben McCarty and Alexis C Garretson and Edwin Taylor and Damien Sileo and Qiuyu Ren and Usman Qazi and Lianghui Li and Jungbae Nam and John B. Wydallis and Pavel Arkhipov and Jack Wei Lun Shi and Aras Bacho and Chris G. Willcocks and Hangrui Cao and Sumeet Motwani and Emily de Oliveira Santos and Johannes Veith and Edward Vendrow and Doru Cojoc and Kengo Zenitani and Joshua Robinson and Longke Tang and Yuqi Li and Joshua Vendrow and Natanael Wildner Fraga and Vladyslav Kuchkin and Andrey Pupasov Maksimov and Pierre Marion and Denis Efremov and Jayson Lynch and Kaiqu Liang and Aleksandar Mikov and Andrew Gritsevskiy and Julien Guillod and Gözdenur Demir and Dakotah Martinez and Ben Pageler and Kevin Zhou and Saeed Soori and Ori Press and Henry Tang and Paolo Rissone and Sean R. Green and Lina Brüssel and Moon Twayana and Aymeric Dieuleveut and Joseph Marvin Imperial and Ameya Prabhu and Jinzhou Yang and Nick Crispino and Arun Rao and Dimitri Zvonkine and Gabriel Loiseau and Mikhail Kalinin and Marco Lukas and Ciprian Manolescu and Nate Stambaugh and Subrata Mishra and Tad Hogg and Carlo Bosio and Brian P Coppola and Julian Salazar and Jaehyeok Jin and Rafael Sayous and Stefan Ivanov and Philippe Schwaller and Shaipranesh Senthilkuma and Andres M Bran and Andres Algaba and Kelsey Van den Houte and Lynn Van Der Sypt and Brecht Verbeken and David Noever and Alexei Kopylov and Benjamin Myklebust and Bikun Li and Lisa Schut and Evgenii Zheltonozhskii and Qiaochu Yuan and Derek Lim and Richard Stanley and Tong Yang and John Maar and Julian Wykowski and Martí Oller and Anmol Sahu and Cesare Giulio Ardito and Yuzheng Hu and Ariel Ghislain Kemogne Kamdoum and Alvin Jin and Tobias Garcia Vilchis and Yuexuan Zu and Martin Lackner and James Koppel and Gongbo Sun and Daniil S. Antonenko and Steffi Chern and Bingchen Zhao and Pierrot Arsene and Joseph M Cavanagh and Daofeng Li and Jiawei Shen and Donato Crisostomi and Wenjin Zhang and Ali Dehghan and Sergey Ivanov and David Perrella and Nurdin Kaparov and Allen Zang and Ilia Sucholutsky and Arina Kharlamova and Daniil Orel and Vladislav Poritski and Shalev Ben-David and Zachary Berger and Parker Whitfill and Michael Foster and Daniel Munro and Linh Ho and Shankar Sivarajan and Dan Bar Hava and Aleksey Kuchkin and David Holmes and Alexandra Rodriguez-Romero and Frank Sommerhage and Anji Zhang and Richard Moat and Keith Schneider and Zakayo Kazibwe and Don Clarke and Dae Hyun Kim and Felipe Meneguitti Dias and Sara Fish and Veit Elser and Tobias Kreiman and Victor Efren Guadarrama Vilchis and Immo Klose and Ujjwala Anantheswaran and Adam Zweiger and Kaivalya Rawal and Jeffery Li and Jeremy Nguyen and Nicolas Daans and Haline Heidinger and Maksim Radionov and Václav Rozhoň and Vincent Ginis and Christian Stump and Niv Cohen and Rafał Poświata and Josef Tkadlec and Alan Goldfarb and Chenguang Wang and Piotr Padlewski and Stanislaw Barzowski and Kyle Montgomery and Ryan Stendall and Jamie Tucker-Foltz and Jack Stade and T. Ryan Rogers and Tom Goertzen and Declan Grabb and Abhishek Shukla and Alan Givré and John Arnold Ambay and Archan Sen and Muhammad Fayez Aziz and Mark H Inlow and Hao He and Ling Zhang and Younesse Kaddar and Ivar Ängquist and Yanxu Chen and Harrison K Wang and Kalyan Ramakrishnan and Elliott Thornley and Antonio Terpin and Hailey Schoelkopf and Eric Zheng and Avishy Carmi and Ethan D. L. Brown and Kelin Zhu and Max Bartolo and Richard Wheeler and Martin Stehberger and Peter Bradshaw and JP Heimonen and Kaustubh Sridhar and Ido Akov and Jennifer Sandlin and Yury Makarychev and Joanna Tam and Hieu Hoang and David M. Cunningham and Vladimir Goryachev and Demosthenes Patramanis and Michael Krause and Andrew Redenti and David Aldous and Jesyin Lai and Shannon Coleman and Jiangnan Xu and Sangwon Lee and Ilias Magoulas and Sandy Zhao and Ning Tang and Michael K. Cohen and Orr Paradise and Jan Hendrik Kirchner and Maksym Ovchynnikov and Jason O. Matos and Adithya Shenoy and Michael Wang and Yuzhou Nie and Anna Sztyber-Betley and Paolo Faraboschi and Robin Riblet and Jonathan Crozier and Shiv Halasyamani and Shreyas Verma and Prashant Joshi and Eli Meril and Ziqiao Ma and Jérémy Andréoletti and Raghav Singhal and Jacob Platnick and Volodymyr Nevirkovets and Luke Basler and Alexander Ivanov and Seri Khoury and Nils Gustafsson and Marco Piccardo and Hamid Mostaghimi and Qijia Chen and Virendra Singh and Tran Quoc Khánh and Paul Rosu and Hannah Szlyk and Zachary Brown and Himanshu Narayan and Aline Menezes and Jonathan Roberts and William Alley and Kunyang Sun and Arkil Patel and Max Lamparth and Anka Reuel and Linwei Xin and Hanmeng Xu and Jacob Loader and Freddie Martin and Zixuan Wang and Andrea Achilleos and Thomas Preu and Tomek Korbak and Ida Bosio and Fereshteh Kazemi and Ziye Chen and Biró Bálint and Eve J. Y. Lo and Jiaqi Wang and Maria Inês S. Nunes and Jeremiah Milbauer and M Saiful Bari and Zihao Wang and Behzad Ansarinejad and Yewen Sun and Stephane Durand and Hossam Elgnainy and Guillaume Douville and Daniel Tordera and George Balabanian and Hew Wolff and Lynna Kvistad and Hsiaoyun Milliron and Ahmad Sakor and Murat Eron and Andrew Favre D. O. and Shailesh Shah and Xiaoxiang Zhou and Firuz Kamalov and Sherwin Abdoli and Tim Santens and Shaul Barkan and Allison Tee and Robin Zhang and Alessandro Tomasiello and G. Bruno De Luca and Shi-Zhuo Looi and Vinh-Kha Le and Noam Kolt and Jiayi Pan and Emma Rodman and Jacob Drori and Carl J Fossum and Niklas Muennighoff and Milind Jagota and Ronak Pradeep and Honglu Fan and Jonathan Eicher and Michael Chen and Kushal Thaman and William Merrill and Moritz Firsching and Carter Harris and Stefan Ciobâcă and Jason Gross and Rohan Pandey and Ilya Gusev and Adam Jones and Shashank Agnihotri and Pavel Zhelnov and Mohammadreza Mofayezi and Alexander Piperski and David K. Zhang and Kostiantyn Dobarskyi and Roman Leventov and Ignat Soroko and Joshua Duersch and Vage Taamazyan and Andrew Ho and Wenjie Ma and William Held and Ruicheng Xian and Armel Randy Zebaze and Mohanad Mohamed and Julian Noah Leser and Michelle X Yuan and Laila Yacar and Johannes Lengler and Katarzyna Olszewska and Claudio Di Fratta and Edson Oliveira and Joseph W. Jackson and Andy Zou and Muthu Chidambaram and Timothy Manik and Hector Haffenden and Dashiell Stander and Ali Dasouqi and Alexander Shen and Bita Golshani and David Stap and Egor Kretov and Mikalai Uzhou and Alina Borisovna Zhidkovskaya and Nick Winter and Miguel Orbegozo Rodriguez and Robert Lauff and Dustin Wehr and Colin Tang and Zaki Hossain and Shaun Phillips and Fortuna Samuele and Fredrik Ekström and Angela Hammon and Oam Patel and Faraz Farhidi and George Medley and Forough Mohammadzadeh and Madellene Peñaflor and Haile Kassahun and Alena Friedrich and Rayner Hernandez Perez and Daniel Pyda and Taom Sakal and Omkar Dhamane and Ali Khajegili Mirabadi and Eric Hallman and Kenchi Okutsu and Mike Battaglia and Mohammad Maghsoudimehrabani and Alon Amit and Dave Hulbert and Roberto Pereira and Simon Weber and Handoko and Anton Peristyy and Stephen Malina and Mustafa Mehkary and Rami Aly and Frank Reidegeld and Anna-Katharina Dick and Cary Friday and Mukhwinder Singh and Hassan Shapourian and Wanyoung Kim and Mariana Costa and Hubeyb Gurdogan and Harsh Kumar and Chiara Ceconello and Chao Zhuang and Haon Park and Micah Carroll and Andrew R. Tawfeek and Stefan Steinerberger and Daattavya Aggarwal and Michael Kirchhof and Linjie Dai and Evan Kim and Johan Ferret and Jainam Shah and Yuzhou Wang and Minghao Yan and Krzysztof Burdzy and Lixin Zhang and Antonio Franca and Diana T. Pham and Kang Yong Loh and Joshua Robinson and Abram Jackson and Paolo Giordano and Philipp Petersen and Adrian Cosma and Jesus Colino and Colin White and Jacob Votava and Vladimir Vinnikov and Ethan Delaney and Petr Spelda and Vit Stritecky and Syed M. Shahid and Jean-Christophe Mourrat and Lavr Vetoshkin and Koen Sponselee and Renas Bacho and Zheng-Xin Yong and Florencia de la Rosa and Nathan Cho and Xiuyu Li and Guillaume Malod and Orion Weller and Guglielmo Albani and Leon Lang and Julien Laurendeau and Dmitry Kazakov and Fatimah Adesanya and Julien Portier and Lawrence Hollom and Victor Souza and Yuchen Anna Zhou and Julien Degorre and Yiğit Yalın and Gbenga Daniel Obikoya and Rai and Filippo Bigi and M. C. Boscá and Oleg Shumar and Kaniuar Bacho and Gabriel Recchia and Mara Popescu and Nikita Shulga and Ngefor Mildred Tanwie and Thomas C. H. Lux and Ben Rank and Colin Ni and Matthew Brooks and Alesia Yakimchyk and Huanxu and Liu and Stefano Cavalleri and Olle Häggström and Emil Verkama and Joshua Newbould and Hans Gundlach and Leonor Brito-Santana and Brian Amaro and Vivek Vajipey and Rynaa Grover and Ting Wang and Yosi Kratish and Wen-Ding Li and Sivakanth Gopi and Andrea Caciolai and Christian Schroeder de Witt and Pablo Hernández-Cámara and Emanuele Rodolà and Jules Robins and Dominic Williamson and Vincent Cheng and Brad Raynor and Hao Qi and Ben Segev and Jingxuan Fan and Sarah Martinson and Erik Y. Wang and Kaylie Hausknecht and Michael P. Brenner and Mao Mao and Christoph Demian and Peyman Kassani and Xinyu Zhang and David Avagian and Eshawn Jessica Scipio and Alon Ragoler and Justin Tan and Blake Sims and Rebeka Plecnik and Aaron Kirtland and Omer Faruk Bodur and D. P. Shinde and Yan Carlos Leyva Labrador and Zahra Adoul and Mohamed Zekry and Ali Karakoc and Tania C. B. Santos and Samir Shamseldeen and Loukmane Karim and Anna Liakhovitskaia and Nate Resman and Nicholas Farina and Juan Carlos Gonzalez and Gabe Maayan and Earth Anderson and Rodrigo De Oliveira Pena and Elizabeth Kelley and Hodjat Mariji and Rasoul Pouriamanesh and Wentao Wu and Ross Finocchio and Ismail Alarab and Joshua Cole and Danyelle Ferreira and Bryan Johnson and Mohammad Safdari and Liangti Dai and Siriphan Arthornthurasuk and Isaac C. McAlister and Alejandro José Moyano and Alexey Pronin and Jing Fan and Angel Ramirez-Trinidad and Yana Malysheva and Daphiny Pottmaier and Omid Taheri and Stanley Stepanic and Samuel Perry and Luke Askew and Raúl Adrián Huerta Rodríguez and Ali M. R. Minissi and Ricardo Lorena and Krishnamurthy Iyer and Arshad Anil Fasiludeen and Ronald Clark and Josh Ducey and Matheus Piza and Maja Somrak and Eric Vergo and Juehang Qin and Benjámin Borbás and Eric Chu and Jack Lindsey and Antoine Jallon and I. M. J. McInnis and Evan Chen and Avi Semler and Luk Gloor and Tej Shah and Marc Carauleanu and Pascal Lauer and Tran Đuc Huy and Hossein Shahrtash and Emilien Duc and Lukas Lewark and Assaf Brown and Samuel Albanie and Brian Weber and Warren S. Vaz and Pierre Clavier and Yiyang Fan and Gabriel Poesia Reis e Silva and Long and Lian and Marcus Abramovitch and Xi Jiang and Sandra Mendoza and Murat Islam and Juan Gonzalez and Vasilios Mavroudis and Justin Xu and Pawan Kumar and Laxman Prasad Goswami and Daniel Bugas and Nasser Heydari and Ferenc Jeanplong and Thorben Jansen and Antonella Pinto and Archimedes Apronti and Abdallah Galal and Ng Ze-An and Ankit Singh and Tong Jiang and Joan of Arc Xavier and Kanu Priya Agarwal and Mohammed Berkani and Gang Zhang and Zhehang Du and Benedito Alves de Oliveira Junior and Dmitry Malishev and Nicolas Remy and Taylor D. Hartman and Tim Tarver and Stephen Mensah and Gautier Abou Loume and Wiktor Morak and Farzad Habibi and Sarah Hoback and Will Cai and Javier Gimenez and Roselynn Grace Montecillo and Jakub Łucki and Russell Campbell and Asankhaya Sharma and Khalida Meer and Shreen Gul and Daniel Espinosa Gonzalez and Xavier Alapont and Alex Hoover and Gunjan Chhablani and Freddie Vargus and Arunim Agarwal and Yibo Jiang and Deepakkumar Patil and David Outevsky and Kevin Joseph Scaria and Rajat Maheshwari and Abdelkader Dendane and Priti Shukla and Ashley Cartwright and Sergei Bogdanov and Niels Mündler and Sören Möller and Luca Arnaboldi and Kunvar Thaman and Muhammad Rehan Siddiqi and Prajvi Saxena and Himanshu Gupta and Tony Fruhauff and Glen Sherman and Mátyás Vincze and Siranut Usawasutsakorn and Dylan Ler and Anil Radhakrishnan and Innocent Enyekwe and Sk Md Salauddin and Jiang Muzhen and Aleksandr Maksapetyan and Vivien Rossbach and Chris Harjadi and Mohsen Bahaloohoreh and Claire Sparrow and Jasdeep Sidhu and Sam Ali and Song Bian and John Lai and Eric Singer and Justine Leon Uro and Greg Bateman and Mohamed Sayed and Ahmed Menshawy and Darling Duclosel and Dario Bezzi and Yashaswini Jain and Ashley Aaron and Murat Tiryakioglu and Sheeshram Siddh and Keith Krenek and Imad Ali Shah and Jun Jin and Scott Creighton and Denis Peskoff and Zienab EL-Wasif and Ragavendran P V and Michael Richmond and Joseph McGowan and Tejal Patwardhan and Hao-Yu Sun and Ting Sun and Nikola Zubić and Samuele Sala and Stephen Ebert and Jean Kaddour and Manuel Schottdorf and Dianzhuo Wang and Gerol Petruzella and Alex Meiburg and Tilen Medved and Ali ElSheikh and S Ashwin Hebbar and Lorenzo Vaquero and Xianjun Yang and Jason Poulos and Vilém Zouhar and Sergey Bogdanik and Mingfang Zhang and Jorge Sanz-Ros and David Anugraha and Yinwei Dai and Anh N. Nhu and Xue Wang and Ali Anil Demircali and Zhibai Jia and Yuyin Zhou and Juncheng Wu and Mike He and Nitin Chandok and Aarush Sinha and Gaoxiang Luo and Long Le and Mickaël Noyé and Michał Perełkiewicz and Ioannis Pantidis and Tianbo Qi and Soham Sachin Purohit and Letitia Parcalabescu and Thai-Hoa Nguyen and Genta Indra Winata and Edoardo M. Ponti and Hanchen Li and Kaustubh Dhole and Jongee Park and Dario Abbondanza and Yuanli Wang and Anupam Nayak and Diogo M. Caetano and Antonio A. W. L. Wong and Maria del Rio-Chanona and Dániel Kondor and Pieter Francois and Ed Chalstrey and Jakob Zsambok and Dan Hoyer and Jenny Reddish and Jakob Hauser and Francisco-Javier Rodrigo-Ginés and Suchandra Datta and Maxwell Shepherd and Thom Kamphuis and Qizheng Zhang and Hyunjun Kim and Ruiji Sun and Jianzhu Yao and Franck Dernoncourt and Satyapriya Krishna and Sina Rismanchian and Bonan Pu and Francesco Pinto and Yingheng Wang and Kumar Shridhar and Kalon J. Overholt and Glib Briia and Hieu Nguyen and David and Soler Bartomeu and Tony CY Pang and Adam Wecker and Yifan Xiong and Fanfei Li and Lukas S. Huber and Joshua Jaeger and Romano De Maddalena and Xing Han Lù and Yuhui Zhang and Claas Beger and Patrick Tser Jern Kon and Sean Li and Vivek Sanker and Ming Yin and Yihao Liang and Xinlu Zhang and Ankit Agrawal and Li S. Yifei and Zechen Zhang and Mu Cai and Yasin Sonmez and Costin Cozianu and Changhao Li and Alex Slen and Shoubin Yu and Hyun Kyu Park and Gabriele Sarti and Marcin Briański and Alessandro Stolfo and Truong An Nguyen and Mike Zhang and Yotam Perlitz and Jose Hernandez-Orallo and Runjia Li and Amin Shabani and Felix Juefei-Xu and Shikhar Dhingra and Orr Zohar and My Chiffon Nguyen and Alexander Pondaven and Abdurrahim Yilmaz and Xuandong Zhao and Chuanyang Jin and Muyan Jiang and Stefan Todoran and Xinyao Han and Jules Kreuer and Brian Rabern and Anna Plassart and Martino Maggetti and Luther Yap and Robert Geirhos and Jonathon Kean and Dingsu Wang and Sina Mollaei and Chenkai Sun and Yifan Yin and Shiqi Wang and Rui Li and Yaowen Chang and Anjiang Wei and Alice Bizeul and Xiaohan Wang and Alexandre Oliveira Arrais and Kushin Mukherjee and Jorge Chamorro-Padial and Jiachen Liu and Xingyu Qu and Junyi Guan and Adam Bouyamourn and Shuyu Wu and Martyna Plomecka and Junda Chen and Mengze Tang and Jiaqi Deng and Shreyas Subramanian and Haocheng Xi and Haoxuan Chen and Weizhi Zhang and Yinuo Ren and Haoqin Tu and Sejong Kim and Yushun Chen and Sara Vera Marjanović and Junwoo Ha and Grzegorz Luczyna and Jeff J. Ma and Zewen Shen and Dawn Song and Cedegao E. Zhang and Zhun Wang and Gaël Gendron and Yunze Xiao and Leo Smucker and Erica Weng and Kwok Hao Lee and Zhe Ye and Stefano Ermon and Ignacio D. Lopez-Miguel and Theo Knights and Anthony Gitter and Namkyu Park and Boyi Wei and Hongzheng Chen and Kunal Pai and Ahmed Elkhanany and Han Lin and Philipp D. Siedler and Jichao Fang and Ritwik Mishra and Károly Zsolnai-Fehér and Xilin Jiang and Shadab Khan and Jun Yuan and Rishab Kumar Jain and Xi Lin and Mike Peterson and Zhe Wang and Aditya Malusare and Maosen Tang and Isha Gupta and Ivan Fosin and Timothy Kang and Barbara Dworakowska and Kazuki Matsumoto and Guangyao Zheng and Gerben Sewuster and Jorge Pretel Villanueva and Ivan Rannev and Igor Chernyavsky and Jiale Chen and Deepayan Banik and Ben Racz and Wenchao Dong and Jianxin Wang and Laila Bashmal and Duarte V. Gonçalves and Wei Hu and Kaushik Bar and Ondrej Bohdal and Atharv Singh Patlan and Shehzaad Dhuliawala and Caroline Geirhos and Julien Wist and Yuval Kansal and Bingsen Chen and Kutay Tire and Atak Talay Yücel and Brandon Christof and Veerupaksh Singla and Zijian Song and Sanxing Chen and Jiaxin Ge and Kaustubh Ponkshe and Isaac Park and Tianneng Shi and Martin Q. Ma and Joshua Mak and Sherwin Lai and Antoine Moulin and Zhuo Cheng and Zhanda Zhu and Ziyi Zhang and Vaidehi Patil and Ketan Jha and Qiutong Men and Jiaxuan Wu and Tianchi Zhang and Bruno Hebling Vieira and Alham Fikri Aji and Jae-Won Chung and Mohammed Mahfoud and Ha Thi Hoang and Marc Sperzel and Wei Hao and Kristof Meding and Sihan Xu and Vassilis Kostakos and Davide Manini and Yueying Liu and Christopher Toukmaji and Jay Paek and Eunmi Yu and Arif Engin Demircali and Zhiyi Sun and Ivan Dewerpe and Hongsen Qin and Roman Pflugfelder and James Bailey and Johnathan Morris and Ville Heilala and Sybille Rosset and Zishun Yu and Peter E. Chen and Woongyeong Yeo and Eeshaan Jain and Ryan Yang and Sreekar Chigurupati and Julia Chernyavsky and Sai Prajwal Reddy and Subhashini Venugopalan and Hunar Batra and Core Francisco Park and Hieu Tran and Guilherme Maximiano and Genghan Zhang and Yizhuo Liang and Hu Shiyu and Rongwu Xu and Rui Pan and Siddharth Suresh and Ziqi Liu and Samaksh Gulati and Songyang Zhang and Peter Turchin and Christopher W. Bartlett and Christopher R. Scotese and Phuong M. Cao and Ben Wu and Jacek Karwowski and Davide Scaramuzza and Aakaash Nattanmai and Gordon McKellips and Anish Cheraku and Asim Suhail and Ethan Luo and Marvin Deng and Jason Luo and Ashley Zhang and Kavin Jindel and Jay Paek and Kasper Halevy and Allen Baranov and Michael Liu and Advaith Avadhanam and David Zhang and Vincent Cheng and Brad Ma and Evan Fu and Liam Do and Joshua Lass and Hubert Yang and Surya Sunkari and Vishruth Bharath and Violet Ai and James Leung and Rishit Agrawal and Alan Zhou and Kevin Chen and Tejas Kalpathi and Ziqi Xu and Gavin Wang and Tyler Xiao and Erik Maung and Sam Lee and Ryan Yang and Roy Yue and Ben Zhao and Julia Yoon and Sunny Sun and Aryan Singh and Ethan Luo and Clark Peng and Tyler Osbey and Taozhi Wang and Daryl Echeazu and Hubert Yang and Timothy Wu and Spandan Patel and Vidhi Kulkarni and Vijaykaarti Sundarapandiyan and Ashley Zhang and Andrew Le and Zafir Nasim and Srikar Yalam and Ritesh Kasamsetty and Soham Samal and Hubert Yang and David Sun and Nihar Shah and Abhijeet Saha and Alex Zhang and Leon Nguyen and Laasya Nagumalli and Kaixin Wang and Alan Zhou and Aidan Wu and Jason Luo and Anwith Telluri and Summer Yue and Alexandr Wang and Dan Hendrycks},
      year={2025},
      eprint={2501.14249},
      archivePrefix={arXiv},
      primaryClass={cs.LG},
      url={https://arxiv.org/abs/2501.14249}, 
}

@misc{han2025deepresearchertesttimediffusion,
      title={Deep Researcher with Test-Time Diffusion}, 
      author={Rujun Han and Yanfei Chen and Zoey CuiZhu and Lesly Miculicich and Guan Sun and Yuanjun Bi and Weiming Wen and Hui Wan and Chunfeng Wen and Solène Maître and George Lee and Vishy Tirumalashetty and Emily Xue and Zizhao Zhang and Salem Haykal and Burak Gokturk and Tomas Pfister and Chen-Yu Lee},
      year={2025},
      eprint={2507.16075},
      archivePrefix={arXiv},
      primaryClass={cs.CL},
      url={https://arxiv.org/abs/2507.16075}, 
}

@misc{li2025websailornavigatingsuperhumanreasoning,
      title={WebSailor: Navigating Super-human Reasoning for Web Agent}, 
      author={Kuan Li and Zhongwang Zhang and Huifeng Yin and Liwen Zhang and Litu Ou and Jialong Wu and Wenbiao Yin and Baixuan Li and Zhengwei Tao and Xinyu Wang and Weizhou Shen and Junkai Zhang and Dingchu Zhang and Xixi Wu and Yong Jiang and Ming Yan and Pengjun Xie and Fei Huang and Jingren Zhou},
      year={2025},
      eprint={2507.02592},
      archivePrefix={arXiv},
      primaryClass={cs.CL},
      url={https://arxiv.org/abs/2507.02592}, 
}

@misc{liu2025webexplorerexploreevolvetraining,
      title={WebExplorer: Explore and Evolve for Training Long-Horizon Web Agents}, 
      author={Junteng Liu and Yunji Li and Chi Zhang and Jingyang Li and Aili Chen and Ke Ji and Weiyu Cheng and Zijia Wu and Chengyu Du and Qidi Xu and Jiayuan Song and Zhengmao Zhu and Wenhu Chen and Pengyu Zhao and Junxian He},
      year={2025},
      eprint={2509.06501},
      archivePrefix={arXiv},
      primaryClass={cs.CL},
      url={https://arxiv.org/abs/2509.06501}, 
}

@misc{mialon2023gaiabenchmarkgeneralai,
      title={GAIA: a benchmark for General AI Assistants}, 
      author={Grégoire Mialon and Clémentine Fourrier and Craig Swift and Thomas Wolf and Yann LeCun and Thomas Scialom},
      year={2023},
      eprint={2311.12983},
      archivePrefix={arXiv},
      primaryClass={cs.CL},
      url={https://arxiv.org/abs/2311.12983}, 
}

\clearpage
\appendix

\section{Appendix}
\label{sec:appendix}

\subsection{Use of large language models (LLMs)}
We acknowledge the use of LLMs (e.g. ChatGPT) for editing the text to correct grammatical errors and improve clarity and flow. All core scientific content and research ideas were authored solely by the authors.

\subsection{Implementation details for SAGE}\label{app:implementation_details}

For data generator and search agent, we set temperature to 1 and disable thinking for gemini-2.5-flash. If the data generator does not generate a ($q$, $a$) pair after issuing the maximum number of search calls, we force it to generate a ($q$, $a$) pair by appending ``\texttt{<think>I have used up all the search budget and I will use the existing information to formulate a new plan and generate the question, answer, and answering plans.}'' to the prompt.

\paragraph{Data generator setting} We include the prompt used for the initial data generator agent in Table \ref{fig:initial_data_generator_prompt}; and the prompt for feedback in Figure \ref{fig:data_generator_correctness_prompt} and Figure \ref{fig:data_generator_difficulty_prompt}.

\paragraph{Search agent setting} We include the prompt used for search agent in Table \ref{fig:search_agent_prompt}.

\paragraph{LLM-as-a-judge setting}

We include the prompt used for reference-based LLM-as-a-judge in Prompt \ref{fig:llm_as_judge_prompt}. We use gemini-2.0-flash with temperature 0.

\begin{figure*}[t]
    \centering
    \caption{Prompt used for initial data generator.}
    \label{fig:initial_data_generator_prompt}
\begin{AIbox}{Prompt used for initial data generator.}
Your task is to generate a complicated question that will require a search agent \texttt{target\_search\_step} search steps to answer by gathering information using a search engine.

You will first reason about the initial document and plan for gathering comprehensive information inside <think> and </think>.

You will then call a search engine by <search> query </search> and it will return the top searched results between <information> and </information> to collect information.

You must conduct reasoning inside <think> and </think> first every time you get new information.

You will call the search engine for \texttt{n\_search\_step} steps.
After \texttt{n\_search\_step} searches, you must provide the question inside <question> and </question>, the answer inside <answer> and </answer>, and the answering step inside <answering steps> and </answering steps>. \
You can use your own knowledge to construct the search query, but the final answer and each of the answering step must be supported by the information you gathered from the search engine.

The question should be understandable standalone as the agent will use the question to search for information without access to the initial document.

An example question: How much did the film in which Jake Gyllenhaal played his second lead role gross in its initial run at the box office?

Avoid How and Why question.

The answer should be \texttt{answer\_type} and short.

Make sure the answer is correct and **unique** for the question generated. 

Initial document: 
\texttt{context}
\end{AIbox}
\end{figure*}

$$$$
\begin{figure*}[t]
    \centering
    \caption{Prompt used for search agent.}
    \label{fig:search_agent_prompt}
\begin{AIbox}{Prompt used for search agent.}
Answer the given question by using a search engine. 

You will first reason about the question inside <think> and </think>, for instance, break down the question into multiple sub-questions that you will search for. 

You must call a search engine by <search> query </search> and it will return the top searched results between <information> and </information>. 

Try to formulate the search query in the form of a question. 

After receiving the information, you must reason about it inside <think> and </think> before issuing a new query or providing the final answer. 

Each of your reasoning step should be grounded in the retrieved information.  Do not use your own knowledge, but you can use commonsense knowledge or arithmetic knowledge. 

Do not use your own knowledge to write the query, the query should be based on the question and the retrieved documents. 

Do not infer the entities in the question, but you can use the entities in the retrieved documents to write the query. 

You can search as many times as your want. Try to break down the question for each search query and gather comprehensive information. 

If you have gathered enough information to answer the question, you can provide the answer to the query inside <answer> and </answer>, without detailed illustrations. 

Generate an answer based on the retrieved information, instead of your own knowledge.

This is an example answer: <answer>Beijing</answer>. Question: \texttt{question} \\
\end{AIbox}
\end{figure*}

\begin{figure*}[t]
    \centering
    \caption{Prompt used for reference-based LLM-as-a-judge.}
    \label{fig:llm_as_judge_prompt}
\begin{AIbox}{Prompt used for reference-based LLM-as-a-judge.}
Judge whether the following [response] to [question] is correct or not based on the precise and unambiguous [correct\_answer\_list] below. Each answer in the [correct\_answer\_list] is separated by a comma.

[question]: \texttt{question}

[response]: \texttt{model\_answer}

Your judgment must be in the format and criteria specified below:

extracted\_final\_answer: The final exact answer extracted from the [response]. Put the extracted answer as 'None' if there is no exact, final answer to extract from the response.

[correct\_answer\_list]: \texttt{gold\_answer}

reasoning: Explain why the extracted\_final\_answer is correct or incorrect based on [correct\_answer\_list], focusing only on if there are meaningful differences between answer in the [correct\_answer\_list] and the extracted\_final\_answer. Focus on recall, i.e. if the extracted\_final\_answer covers all the points in the answer in the [correct\_answer\_list]. It is ok if it provides more details. 

It is also ok if the extracted\_final\_answer misses minor point from the correct\_answer, as long as it is evident that they are referring to the same thing. Do not comment on any background to the problem, do not attempt to solve the problem, do not argue for any answer different than [correct\_answer\_list], focus only on whether the answers match. Ignore capitalization.

correct: Answer 'yes' if extracted\_final\_answer matches any of the answers in [correct\_answer\_list] given above, or is within a small margin of error for numerical problems. Answer 'no' otherwise, i.e. if there is any inconsistency, ambiguity, non-equivalency, or if the extracted answer is incorrect.

confidence: The extracted confidence score between 0\% and 100\% from [response]. Put 100 if there is no confidence score available.
\end{AIbox}
\end{figure*}

\begin{figure*}[t]
    \centering
    \caption{Prompt used for incorporating execution feedback to update the incorrect QA pair.}
    \label{fig:data_generator_correctness_prompt}
\begin{AIbox}{Prompt used for incorporating execution feedback to update the incorrect QA pair.}
 You will be given an output from a question generator agent, which generates a complicated question, answer pair; as well as the output from a search agent, which attempts to solve the question generated in a fixed number of turns. 

The answer from the search agent is not the same as the data generator agent. You task is to examine their traces and output the correct question, answer pair based on their retrieved documents. You can update either the question, the answer or both.

You will first reason about why is there a discrepancy between the search agent's answer and the data generator's answer. Output your reasoning trace inside <reason> and </reason>.
You will then reason about how to update the question answer pair to make sure it is correct and requires the agent \texttt{target\_step} search step to answer. A search step is defined as a call to the search tool. Output your reasoning trace inside <think> and </think>.
For factual information, you should ONLY rely on the context provided for the data generator agent and the documents retrieved by both the data generator and search agent (inside <information> and </information>).

If you find it non-trivial to update just the question and answer, you can generate a new question answer pair ONLY based on the retrieved documents.

The updated question should require the search agent at least \texttt{target\_step} search steps to answer. However, the answer should be short, such as an entity, a date or a number. The question should be understandable standalone, as the search agent will solve the question without access to the documents (they will need to search for them).

When you are ready to provide the new question, answer pair, you can provide the question inside <question> and </question>, the answer inside <answer> and </answer>, and the search step inside <search steps> and </search steps>. 
For each search step, output the exact search question; the sub-answer to the search question; and the retrieved document from the search agent and data generator agent's output that supports the sub-answer. 
Make sure each step is absolutely needed to answer the question and there is no short cut. Tip: use retrieved document from different steps so avoid two sub-queries being solved by one search query.

\# Data generator agent

Prompt:  \texttt{data\_generator\_agent\_prompt}

Agent's output:  \texttt{data\_generator\_agent\_response}

\# Search agent

Prompt: \texttt{search\_agent\_prompt}

Agent's output: \texttt{search\_agent\_response}

\# Your output \\
\end{AIbox}
\end{figure*}

\begin{figure*}[t]
    \centering
    \caption{Prompt used for incorporating execution feedback to update the easy QA pair.}
    \label{fig:data_generator_difficulty_prompt}
\begin{AIbox}{Prompt used for incorporating execution feedback to update the easy QA pair.}
You will be given an output from a question generator agent, which generates a complicated question, answer pair to be solved by a search agent for at least \texttt{target\_step} **search** steps; as well as the output from a search agent, which attempts to solve the question generated. The search agent is able to solve the question in less than\texttt{target\_step} search steps. Your task is to update the question so that it requires the search agent more steps to solve.

You will first reason about why the search agent is able to solve the question in fewer steps. Output your reasoning trace inside <reason> and </reason>.
You will then reason about how to update the question so that it will require more search steps.
For factual information, you should ONLY rely on the context provided for the data generator agent and the documents retrieved by both the data generator and search agent (inside <information> and </information>), without relying on other information not in the retrieved context.
Output your reasoning trace inside <think> and </think>.
If you find it non-trivial to update the plan, you can generate a new question answer pair ONLY based on the retrieved documents.

The updated question should require the search agent at least \texttt{target\_step} search steps to answer. Note that some of the answering steps do not involve search and thus do not count. However, the answer should be short, such as an entity, a date or a number. The question should be understandable standalone, as the agent will solve the question without access to the documents (they will need to search for them).

When you are ready to provide the new question, answer pair, you can provide the question inside <question> and </question>, the answer inside <answer> and </answer>, and the search step inside <search steps> and </search steps>. For each search step, output the exact search question; the sub-answer to the search question; and the retrieved document from the search agent and data generator agent's output that supports the sub-answer. Make sure each step is absolutely needed to answer the question and there is no short cut. Tip: use retrieved document from different steps so avoid two sub-queries being solved by one search query.

\# Data generator agent

Prompt:  \texttt{data\_generator\_agent\_prompt}

Agent's output:  \texttt{data\_generator\_agent\_response}

\# Search agent

Prompt: \texttt{search\_agent\_prompt}

Agent's output: \texttt{search\_agent\_response}

\# Your output \\
\end{AIbox}
\end{figure*}

\subsection{Implementation details for training the search agent}\label{app:implementation_details_search_agent}

\paragraph{PPO objective.} We conduct downstream evaluation by training search agents with the Proximal Policy Optimization (PPO)~\citep{schulman2017proximal}. It optimizes the language model by maximizing the below objective in Figure \ref{alg:ppo}, with $\pi_{theta}$ and $\pi_{old}$ representing the current the previous policy model; $I(y_{t})$ representing whether the loss masking apply to the token, we apply loss masking to retrieved document ($I(y_{t})=0$). $\varepsilon$ is a clipping hyperparameter, and the advantage estimate $A_{t}$ is computed using Generalized Advantage Estimation~\citep{Schulman2015HighDimensionalCC}.

\paragraph{Training details.} For training the search agent, we set the learning rate of the policy LLM to 1e-6 and that of the value LLM to 1e-5. Training is conducted for 500 steps, with warm-up ratios of 0.285 and 0.015 for the policy and value models, respectively. We use Generalized Advantage Estimation (GAE) with parameters $\lambda$ = 1 and $\gamma$ = 1. Training is conducted on a single node with 8 H100 GPUs. We use a total batch size of 512, with a mini-batch size of 256 and a micro-batch size of 64. The maximum sequence length is set to 8,192 tokens, with a maximum response length of 1024 and a maximum length of 1000 tokens for retrieved content. We set the maximum number of retrieval calls to 8.

\begin{figure*}
\small
\[
\mathcal{J}_{\rm PPO}(\theta)
= \mathbb{E}_{x \sim \mathcal{D},\, y \sim \pi_{\rm old}(\cdot \mid x; \mathcal{R})}
\left[
\frac{1}{\sum_{t=1}^{|y|} I(y_t)} \sum_{t=1 : I(y_t)=1}^{|y|}
\min\!\Bigl(
\frac{\pi_\theta(y_t \mid x, y_{<t}; \mathcal{R})}{\pi_{\rm old}(y_t \mid x, y_{<t}; \mathcal{R})} \, A_t,\;
\mathrm{clip}\bigl(r_t(\theta),\,1-\varepsilon,\,1+\varepsilon\bigr)\, A_t
\Bigr)
\right]
\]
\caption{PPO objective for training search agents.}\label{alg:ppo}
\end{figure*}

\subsection{Analysis details}\label{app:analysis_details}
We include the prompt for reasoning strategy analysis in Prompt \ref{fig:reasoning_strategy_prompt}, which also presents the definition for each of the category.

\begin{figure*}[t]
    \centering
    \caption{Prompt used for reasoning strategy analysis.}
    \label{fig:reasoning_strategy_prompt}
\begin{AIbox}{Prompt used for reasoning strategy analysis.}
You will be given a question and a trace of a search agent solving this question. You will analyze and categorize the behavior for each of the thinking step.

Below are some example categories, please feel free to propose new ones as you see appropriate:

- Information inference: the agent makes an inference based on the piece of information in the retrieved document in its reasoning.

- Conflict resolution: the agent reasons about conflicting information in the documents and makes a decision.

- Calculation: The agent performs numerical calculation.

- Temporal reasoning: The agent performs temporal reasoning, such as deriving duration between two dates.

- Self-correction: The agent recognizes that the previous search failed to yield the required information (a list of stations). It re-evaluates its state, confirms its goal, and decides to try a more specific search query.

- Hypothesis Generation: The agent makes a guess / hypothesis that is not grounded in the retrieved documents.

Output format: only return the list of strategies for each step:

- Step i: [list of strategies]

- Step i+1: [list of strategies]

Question: \texttt{question}
Agent's reasoning trace: \texttt{agent\_trace}
\end{AIbox}
\end{figure*}

\begin{table*}[ht!]
\small
\begin{tabular}{p{15cm}}
\toprule
 \textbf{Example questions} \\
\midrule
 \textit{Previous question:} On what date did Tottenham Hotspur win their first major trophy, a feat that made them the only non-league club to achieve it, while under the chairmanship of the individual who took office the year before the club's establishment at its original ground, colloquially known as "The Lane"? 

\textit{Previous answer:} April 27, 1901

\textit{Updated question: } What was the date of the replay match in which the player, who notably scored goals in every round of Tottenham Hotspur's first major trophy victory, helped his team secure that trophy after the club had become a limited company under a chairman who served until 1943?

\textit{Updated answer:} April 27, 1901 \\ 

\midrule

 \textit{Previous question: } What is the death date of the actor who sang songs in the 1944 film that was Naushad's first musical success?  

\textit{Previous answer: } 2 August 1979

\textit{Updated question: } What is the death date of the actor, born in Gujranwala, who sang songs in Naushad's first musical success film of 1944?

\textit{Updated answer:} 2 August 1979 \\ 

\midrule

 \textit{Previous question:} What was the precise date Sir Frank Whittle, a key figure honored at the Midlands Air Museum, received his knighthood as Knight Commander of the Order of the British Empire (KBE) for his pioneering work on the jet engine? 

 \textit{Previous answer:} July 1, 1948 

 \textit{Updated question: } What was the precise date of the military retirement of the figure whose heritage center is located within the museum adjacent to the former Electric Railway Museum site?

\textit{Updated answer:} 26 August 1948 \\

\bottomrule
\end{tabular}
\caption{Example question generated and updated by our pipeline.}
\label{tab:example_question}
\end{table*}

\begin{table*}
\begin{center}
\footnotesize
\begin{tabular}{@{}lrrr@{}}
\toprule
\textbf{Name} & \textbf{Retriever} & \textbf{Explicit structure} & \textbf{Input} \\
\midrule
WebDancer~\citep{webdancer} & Google Search & Inter-page hyperlinks &  Web page or seed entity \\
WebShaper~\citep{webshaper} & Google Search & Knowledge graph and inter-page hyperlinks & Webpage \\
WebSailor~\citep{li2025websailornavigatingsuperhumanreasoning} & Google Search & Knowledge graph & Seed entity \\ 
WebPuzzle~\citep{pangu-deepdiver} & Web Search & Pre-selected rare entities & Web page or seed entity \\
WebExplorer~\citep{liu2025webexplorerexploreevolvetraining} & Google Search & No & Seed entity \\
\midrule
\texttt{SAGE} (Ours) & Wikipedia-based & No & Passage \\
\bottomrule
\end{tabular} 
\end{center}
\caption{Comparison with concurrent work on synthetic data generation for search agents. “Explicit structure” refers to reliance on predefined or constructed graphs, entity links, or formal schemas during data generation. None of the concurrent work open-sources large-scale training data.}
\label{tab:concurrent_work_comparison}
\end{table*}

\end{document}